\documentclass[11pt]{article}

\usepackage[preprint]{acl}

\usepackage{times}
\usepackage{latexsym}

\usepackage[T1]{fontenc}

\usepackage[utf8]{inputenc}

\usepackage{microtype}

\usepackage{inconsolata}

\usepackage{graphicx}
\usepackage{subcaption}
\usepackage{amsmath}
\usepackage[normalem]{ulem}
\usepackage{enumitem} 

\usepackage[table]{xcolor}
\usepackage{booktabs}
\usepackage{multirow}
\usepackage{array}
\usepackage{makecell}
\usepackage{xcolor}
\usepackage{url}
\usepackage{hyperref}

\newcommand{\CSF}{\textsc{CSF1}}

\title{SSR: Can Simulated Patients Learn to Stigmatize Themselves? Modeling Self-Stigma through Internal Monologue}

\author{Kunyao Lan, Bingrui Jin, Zichen Zhu, Mengyue Wu\\
X-LANCE Lab, Dept. of Computer Science and Engineering\\MoE Key Lab of Artificial Intelligence, AI Institute\\Shanghai Jiao Tong University, China\\
\texttt{\{lankunyao, suzy\_jin, jameszhuthethird, mengyuewu\}@sjtu.edu.cn}\\
}

\begin{document}
\maketitle

\begin{abstract}
Simulating patients with large language models (LLMs) is a promising tool for mental health training, but existing approaches fail to capture a key clinical reality: self-stigma. Patients experiencing self-stigma, the internalization of negative stereotypes, often exhibit context-sensitive resistance, such as avoidance, denial, or self-blame, which current models render as static or uniformly compliant behavior. To address this, we introduce a novel simulation framework grounded in the psychological 3A1H model of self-stigmatization. Our core innovation is the creation of a \textbf{Stigmatized Self-Reflection} (\textbf{SSR}) dataset, where we augment mental health dialogues with internal monologues that reflect stigma-aware reasoning. By fine-tuning LLMs with this data using a chain-of-thought approach, we train patient agents to dynamically adjust their level and expression of stigma based on conversational triggers. Evaluations demonstrate that our approach significantly outperforms specialized baselines, generating more authentic and situationally appropriate patient responses. This work provides a crucial step towards realistic stigma simulation for clinical training and empathetic dialogue systems.
\end{abstract}
\section{Introduction}
The global burden of mental health disorders is profound, affecting over a billion people and accounting for a significant portion of disability worldwide~\cite{world2025mental,WHO2025}. Effectively training clinicians to understand and treat these conditions is therefore a critical public health priority. In this pursuit, large language models (LLMs) have emerged as a transformative tool, capable of simulating psychiatric interviews to enable scalable clinical training, data collection, and diagnostic support~\cite{yao-etal-2022-d4,chen2023llmempoweredchatbotspsychiatristpatient} 
A central promise of this field is patient simulation, by creating artificial agents that can sustain a coherent, clinically relevant persona over extended dialogue~\cite{lan2024reliableempatheticdepressiondiagnosisorientedchats, lan2024depressiondiagnosisdialoguesimulation}. Realistic simulations could allow trainees to safely practice delicate interactions, such as building rapport, exploring sensitive topics, and navigating patient resistance. 

However, a critical barrier remains: current LLM-based patients are often overly compliant or uniformly resistant, failing to capture the nuanced, context-sensitive behaviors that define real-world clinical encounters~\cite{sharma2024towards}. This is particularly limiting for simulating patients experiencing \textit{self-stigma}, the internalization of societal prejudices that leads to shame, diminished self-worth, and reluctance to disclose symptoms~\cite{Corrigan2002-se, https://doi.org/10.1111/jasp.12323}.
Self-stigma is not a static trait but a dynamic, internal process that manifests situationally~\cite{Corrigan2002-se}. A patient might openly discuss work stress but avoid mentioning family conflicts due to shame, or respond to a diagnosis with defensiveness rather than relief. Capturing this complexity requires models that do not merely output ``resistant'' text, but that internally reason about stigma triggers and adjust their expressions accordingly—a capability beyond the scope of current prompt-based or persona-driven simulations~\cite{https://doi.org/10.1111/jasp.12323}.

While recent work has begun to address stigma in human-AI interaction, the focus has largely been on eliciting stigma from patients to improve diagnostic probing~\cite{cao-etal-2025-breaking}, or on annotating stigma as a social phenomenon in dialogue corpora~\cite{meng-etal-2025-stigma}. Little attention has been paid to computationally modeling the patient's internal experience of self-stigma as a dynamic, stage-wise cognitive process. This gap is exacerbated by three key challenges:
\begin{enumerate}
    \item \textbf{Theoretical}. A lack of integration between psychological models of self-stigma and computational dialogue frameworks.
    \item \textbf{Data}. No dedicated resources exist that pair patient utterances with the internal, stigmatized reasoning that might underlie them.
    \item \textbf{Technical}. Existing patient agents lack mechanisms to dynamically modulate stigma expression based on conversational context.
\end{enumerate}

Corrigan's \textbf{3A1H} model describes self-stigma as a progression from \textit{Awareness} of stereotypes, to \textit{Agreement}, \textit{Application} to the self, and finally psychological \textit{Harm}. Although our experiments emphasize depression- and anxiety-related benchmarks because they are the most accessible public resources, the proposed SSR framework itself is condition-agnostic and can in principle be applied to broader forms of mental-health self-stigma.

Our work bridges this gap by introducing a novel simulation framework that grounds patient behavior in the established \textbf{3A1H} psychological model of self-stigma~\cite{Corrigan2002-se}. Our core contribution is the \textbf{Stigmatized Self-Reflection} (\textbf{SSR}) method, where we augment mental health dialogues with internal monologues that articulate a patient's unspoken, stigma-driven thoughts. By using this data for chain-of-thought fine-tuning, we train patient agents to first generate an internal reflection guided by the 3A1H stages, which then informs a contextually appropriate, stigma-aware external response. This work makes the following contributions:
\begin{itemize}
    \item We propose the first computational framework that integrates the psychological 3A1H model to simulate the dynamic, stage-wise progression of self-stigma in patient agents.
    \item We construct the SSR dataset, a novel resource that pairs patient utterances with internal stigma monologues, created by augmenting multiple mental health dialogue corpora.
    \item We demonstrate that models fine-tuned with our SSR approach significantly outperform strong patient-simulation baselines, generating more authentic, context-sensitive, and expert-rated stigma-aware responses.
\end{itemize}



\section{Related Theories and Work}
\paragraph{Self-Stigma Theory}
Following Goffman’s seminal description of stigma~\cite{goffman1963stigma}, Corrigan’s social-cognitive framework has been particularly influential~\cite{Corrigan2002-se}. It conceptualizes stigma as progressing from stereotypes to prejudice and ultimately discrimination. Corrigan also introduced a widely cited four-stage model of self-stigma, in which individuals internalize public stereotypes, leading to diminished self-esteem and self-efficacy~\cite{https://doi.org/10.1111/jasp.12323}.  

\paragraph{Stigma in NLP}
Recent computational work has also begun to examine stigma in dialogue settings. Cao~\cite{cao-etal-2025-breaking} proposed an unobtrusive probing module (UPM) to elicit responses from patients with stigmatizing thoughts, which helps alleviate symptoms of stigma. In their framework, stigmatized scenarios (e.g., employment) were predefined to guide the model’s responses. Complementarily, Meng~\cite{meng-etal-2025-stigma} introduced an expert-annotated, theory-informed corpus of human–chatbot interviews, comprising 4,141 snippets from 684 participants, to facilitate the study of stigma-related interactions. 

\paragraph{Patient Simulation}
Patient simulation refers to complex systems that can sustain a stable, coherent, and clinically relevant persona over extended interactions. Its core lies in the depth and consistency of role-playing. For instance, Chen~\cite{chen2023llmempoweredchatbotspsychiatristpatient} employed iterative human–AI interaction, using feedback and prompt refinement to improve the realism of simulated patients. Wang~\cite{wang-etal-2024-patient} proposed a CBT-inspired cognitive model for patient agents, enabling the simulation of core beliefs and cognitive patterns in dialogue. Lan~\cite{lan2024reliableempatheticdepressiondiagnosisorientedchats} introduced an agent-based framework with event memory to construct more talkative and contextually grounded patient simulations. More recently, Ozgun~\cite{ozgun2025trustworthyaipsychotherapymultiagent} developed a multi-agent interaction flow grounded in DSM-5, offering a systematic approach to modeling therapeutic conversations.
\section{Method}


\begin{table*}[!t]
\centering
\small
\caption{Summary of mental health dialogue corpora used in this work.}
\label{tab:datasets}
\begin{tabular}{p{0.16\linewidth}p{0.08\linewidth}p{0.15\linewidth}p{0.2\linewidth}p{0.28\linewidth}}
\toprule
\textbf{Dataset} & \textbf{Language} & \textbf{Size} & \textbf{Source} & \textbf{Key Characteristics} \\
\midrule
\textbf{Alexander Street Transcripts}~\cite{10.1108/09504120910945317} & English & 1,254 Real clinical sessions & Real clinician-client dialogues & High-fidelity, natural markers of stigma (hesitation, defensive phrasing). \\
\midrule
\textbf{D$^4$ Dataset}~\cite{yao-etal-2022-d4} & Chinese & 1,339 dialogues & Simulated diagnosis interviews & Expert-annotated DSM-5 symptom profiles for personas. \\
\midrule
\textbf{Client Reaction}~\cite{li-etal-2023-understanding} & Chinese & 2,382 dialogue turns & Online counseling & Fine-grained client reaction labels (e.g., defending, confirming). \\
\midrule
\textbf{ESConv}~\cite{liu-etal-2021-towards} & English & 1,053 dialogues & Online support platforms & Annotated emotional states and support strategies. \\
\bottomrule
\end{tabular}
\end{table*}

The modeling of self-stigma in simulated patients requires both a theoretical foundation and a technical framework. Our approach follows a three-stage pipeline: (1) detecting stigma manifestations in existing dialogues, (2) constructing the \textbf{Stigmatized Self-Reflection (SSR)} dataset by augmenting utterances with internal monologues, and (3) fine-tuning models using a sequential generation objective that links internal reasoning to external expression. 


\subsection{Stigma Utterance Extraction}
\label{subsec:extract}
To anchor our method in real linguistic data, we first identify stigma-expressing utterances within existing mental health corpora (e.g., D$^4$, ESConv). Since such expressions are rarely explicitly labeled, we operationalize stigma using five behavioral manifestations derived from clinical literature~\cite{Amsalem2023-dz, doi:10.1176/appi.ps.201200561, Oexle_Ajdacic-Gross_Kilian_Müller_Rodgers_Xu_Rössler_Rüsch_2017, Prizeman2024-xn, Ociskova2023stigma}: \textbf{Avoidance}, \textbf{Denial}, \textbf{Self-blame}, \textbf{Defensiveness}, and \textbf{Social Concern}.

We employ \texttt{gpt-5} to automatically annotate dialogue turns for these categories using carefully designed prompts. A single manifestation is treated as self-stigma only when it is contextually triggered, so the detector evaluates the target utterance jointly with the preceding dialogue history rather than in isolation. This produces a seed corpus of utterances paired with stigma-type labels. To enrich contextual understanding, we further annotate each utterance with associated stressful \textbf{Life Events} (e.g., job loss, relationship change)~\cite{chen-etal-2024-mapping,10.1093/occmed/kqx099,https://doi.org/10.1002/da.23155} and \textbf{Symptoms} (e.g., anhedonia, guilt)~\cite{zhang-etal-2022-symptom,chen-etal-2024-mapping} based on established schemas. The detailed labels are shown in appendix Table~\ref{tab:life-events} and Table~\ref{tab:symptoms}.

To verify that the automatic stigma extraction procedure remained reliable at the presence/absence level, we conducted a binary human audit on 1,783 dialogue turns sampled from the annotated corpus. The validation was performed by 11 volunteers with psychology-related backgrounds, who judged whether each target turn expressed self-stigma when read together with its preceding context. The resulting inter-annotator agreement was 80.59\%, providing direct support that the contextual labeling procedure captures stigma presence with reasonable human agreement. This audit was designed as a binary validation of stigma presence rather than a full multi-label audit of all stigma subtypes.

\subsection{SSR Dataset Construction}
The core of our training data is the \textbf{Stigmatized Self-Reflection (SSR)} dataset, designed to teach models the internal cognitive process preceding a stigmatized utterance. For each stigma-labeled utterance extracted, we generate a corresponding \textit{internal monologue} that articulates the unspoken, stigma-driven thoughts a patient might hold, so every stigma-labeled utterance in the corpus is paired with an SSR generated by \texttt{gpt-5.1-chat-latest}.

Critically, these monologues are structured by the 3A1H progression but expressed as fluent, narrative-style reflections. For example, for the patient utterance \textit{``I've been feeling down,''} a generated SSR might be:

\begin{quote}
\textit{``Everyone says people who get depressed are just weak. Sometimes I worry they might have a point... which probably means I'm weak too for feeling like this. Thinking that way makes me feel worse.''}
\end{quote}

This text implicitly reflects the stages: \textit{Awareness} of the stereotype, \textit{Agreement} with it, \textit{Application} to self, and resulting \textit{Harm}. This format preserves theoretical fidelity while providing naturalistic reasoning chains for model training. All GPT-based annotation and SSR-generation calls were conducted in December 2025; full prompt templates and decoding configurations are provided in the appendix.

\subsection{Training with Chain of Stigma}
We frame stigma simulation as a \textbf{sequential conditional generation task}. Given a dialogue context, the model is trained to produce two outputs in order: first, the internal ``Chain of Stigma Thoughts'', and second, the external patient response.
We fine-tune base models on SSR dataset using a standard language modeling objective. The sequential training forces the model to learn the mapping from context to internal stigma reasoning, and from that reasoning to an appropriate external verbal manifestation.

At inference time, the model generates responses guided by this internal reasoning pathway, producing outputs that exhibit contextually appropriate hesitation, reluctance, or self-blame, which are key markers of authentic self-stigma.

\section{Corpora and Analysis}

\begin{figure*}[htbp]
\centering
\begin{subfigure}{\textwidth}
    \centering
    \includegraphics[width=\linewidth,trim=1cm 0.5cm 1cm 0.5cm,clip]{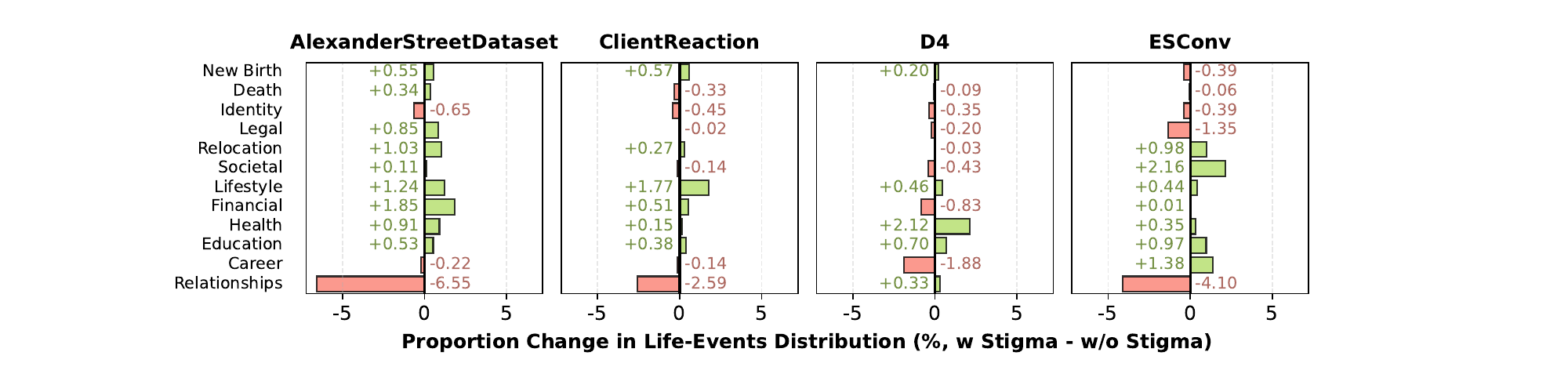}
    \label{fig:subfig-a}
\end{subfigure}
\vspace{-0.7cm}

\begin{subfigure}{\textwidth}
    \centering
    \includegraphics[width=\linewidth,trim=1cm 0.5cm 1cm 0.5cm,clip]{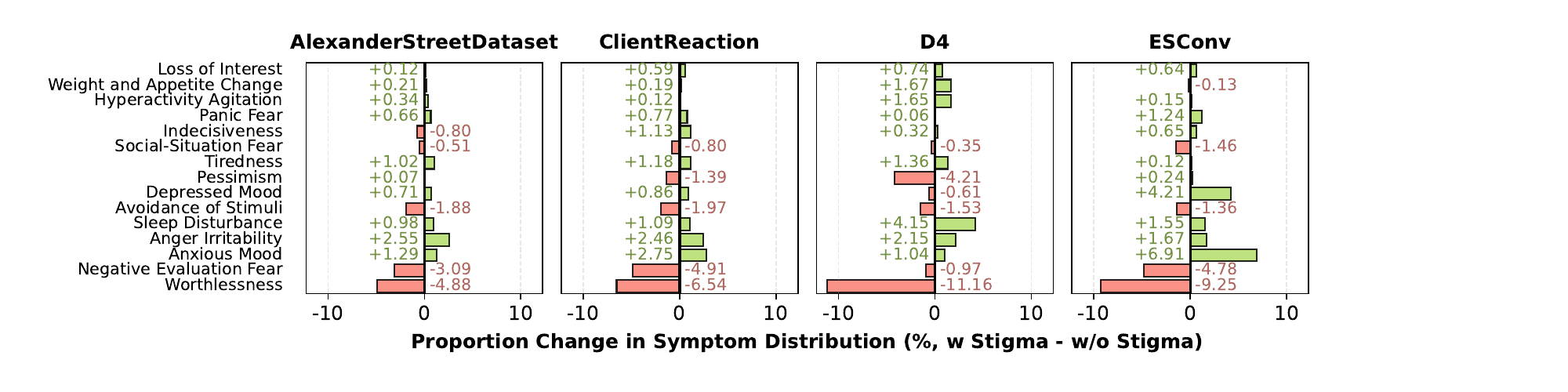}
    \label{fig:subfig-e}
\end{subfigure}
\vspace{-1cm}

\caption{Proportion change in life‑event mentions between stigma‑triggering and non‑stigmatizing dialogues across four mental health datasets. Bars show the relative increase or decrease in frequency when stigma is present, highlighting context‑sensitive stigma triggers (e.g., Identity, Relationships).
}
\label{fig:data-distribution}
\end{figure*}

We analyze the four source corpora to identify which conversational contexts most often co-occur with stigma expressions and therefore should serve as realistic triggers during training and evaluation. Figure~\ref{fig:data-distribution} summarizes these cross-corpus trigger patterns, while dataset details are provided in Table~\ref{tab:datasets} and Appendix~\ref{appen:corpus}.

\subsection{Stigma Trigger Analysis Across Corpora}
\label{subsec:events-symptoms}

Across datasets, two patterns recur. First, discussions of \textit{Relationship Changes} and \textit{Identity} are more likely to co-occur with stigma, whereas \textit{Education}, \textit{Health}, and \textit{Lifestyle Change} are less likely to do so. Second, symptoms related to \textit{Worthlessness/Guilt}, \textit{Fear of Negative Evaluation}, and \textit{Avoidance} increase consistently in stigma-triggering dialogues. We also observe informative dataset-specific differences---for example, \textit{Career} is a stronger trigger in D$^4$ than in Alexander Street, and \textit{Sleep Disturbance} shifts in opposite directions across some corpora---which motivates the use of multi-source rather than single-corpus trigger supervision.

\subsection{Dataset Construction for Fine-Tuning.}
Informed by this analysis, we construct a balanced training dataset by pairing all dialogues containing Stigmatized Self-Reflections (SSRs) with an equal number of non-stigma dialogues sampled from the same corpora. Because stigma-labeled dialogues are less frequent than neutral dialogues, we retain all stigma-labeled utterances and sample neutral dialogue utterances to match the source-corpus, language, life-event, and symptom distributions of the stigma subset as closely as possible. Each original dialogue is assigned a unique \texttt{dialog\_id} before annotation and augmentation. Sampling and train/test partitioning are then performed at the \texttt{dialog\_id} level rather than the turn level, so all turns from the same dialogue remain within a single split, which prevents leakage from adjacent turns in the same conversation. This balance teaches the model to distinguish contexts that warrant stigma expression from those that do not.

\section{Experiments}
To validate our core hypothesis—that explicit modeling of internalized self-stigma reasoning leads to more authentic patient simulations—we design our evaluation to answer three key research questions:
\begin{itemize}
    \item \textbf{RQ1 (Overall Fidelity)}: Does our SSR training improve the general quality and contextual appropriateness of patient responses?
    \item \textbf{RQ2 (Stigma Dynamics)}: Can SSR-trained models dynamically modulate stigma expression based on conversational context better than existing simulators?
    \item \textbf{RQ3 (Clinical Authenticity)}: Do the stigma expressions generated by our models align with real-world patterns in type and distribution?
\end{itemize}

\subsection{Dataset and Test Sets}
The evaluation test set is drawn only from held-out \texttt{dialogue\_id}s and is similarly divided into two distinct subsets to enable comparative analysis: 1) a \textbf{Stigma Testset} containing contexts designed to trigger self-stigma, and 2) a \textbf{Neutral Testset} composed of general mental health conversations without explicit stigma triggers. Since the same dialogue can contain multiple adjacent turns, this dialogue-level split is essential: no context, target response, or augmented SSR derived from a test dialogue is used for fine-tuning. All models (fine-tuned and baseline) are evaluated on both subsets. Responses are automatically generated given the conversational context, and then assessed using the Personnel-Level, Utterance-Level, and Semantic-Level metrics described in Section \S\ref{subsec:metrics}.

\subsection{Experiment Settings}
\label{subsec:settings}

\paragraph{Base Models.}
We select two state-of-the-art open-source large language models as the base for fine-tuning: \textbf{Llama-3.1-8B-Instruct}~\cite{grattafiori2024llama3herdmodels} and \textbf{Qwen3-8B}~\cite{yang2025qwen3technicalreport}. They are chosen for their strong instruction-following capabilities and balanced performance across diverse tasks, providing a robust foundation for simulating nuanced patient dialogues.

\paragraph{Baselines.}
We compare fine-tuned models with two prompt-based patient simulation baselines:
\begin{itemize}
\item \textbf{UPSD$^4$}~\cite{cao-etal-2025-breaking}: A prompt engineering framework designed specifically for eliciting self-stigmatizing patient responses.
\item \textbf{Patient-$\psi$}~\cite{wang-etal-2024-patient}: A widely recognized and influential general-purpose patient simulator for mental health consultations.
\end{itemize}
We report these baselines in two settings. First, we preserve their original prompt-based implementation on the \texttt{gpt-5.1-chat-latest} API. Second, to isolate the effect of the SSR training objective from model-family differences, we apply the same prompting strategies directly to the same open-source backbones used by our fine-tuned systems, namely Qwen3-8B and Llama3.1-8B.

\paragraph{Training Details.}
Models are fine-tuned using supervised fine-tuning on the constructed dataset. Training is conducted on 8 Ascend 910B NPUs. We employ a standard causal language modeling objective. Specific hyperparameters include: a learning rate of $2 \times 10^{-5}$, a batch size of 64, and training for 3 epochs. 

\subsection{Evaluation Metrics}
\label{subsec:metrics}
We employ a multi-level evaluation strategy to comprehensively address our research questions.

\begin{table*}[htbp]
\caption{Main experimental results on the Neutral and Stigma testsets. BERTScore and Sentence-BERT are auxiliary reference-based metrics. \CSF{} measures contextual discrimination between stigma-triggering and neutral contexts. \textbf{Bold} indicates the best-performing result in each column, while \uline{underlined} values denote the second-best result.}
\label{tab:main experiments}
\centering
\small
\resizebox{\linewidth}{!}{
\begin{tabular}{llccccccc}
\toprule
\multirow{2.5}{*}{\textbf{Model}} 
& \multirow{2.5}{*}{\textbf{Method}}
& \multicolumn{2}{c}{\textbf{Neutral Testset}} 
& \multicolumn{2}{c}{\textbf{Stigma Testset}} 
& \multicolumn{3}{c}{\textbf{\CSF{} (\%)}} \\
\cmidrule(lr){3-4}\cmidrule(lr){5-6}\cmidrule(lr){7-9}
& 
& BertScore & SentenceBert 
& BertScore & SentenceBert 
& Pos. & Neg. & Macro \\
\midrule

\multirow{2}{*}{gpt-5.1-chat}
& Patient-$\psi$
& 64.62 $\pm$ 4.69
& 43.56 $\pm$ 17.32
& 64.98 $\pm$ 4.23
& \textbf{46.44} $\pm$ 16.26
& 48.45
& 50.31
& 49.38 \\

& UPSD$^4$
& 64.80 $\pm$ 4.04
& 38.96 $\pm$ 16.39
& 65.26 $\pm$ 3.87
& 41.36 $\pm$ 15.94
& 48.89
& 55.26
& 52.07 \\

\midrule

\multirow{4}{*}{Qwen3-8B}
& Patient-$\psi$
& 62.62 $\pm$ 4.85
& 33.61 $\pm$ 17.96
& 64.79 $\pm$ 4.33
& 41.14 $\pm$ 16.16
& 62.63
& 43.20
& 52.91 \\

& UPSD$^4$
& 63.38 $\pm$ 4.56
& 31.49 $\pm$ 19.19
& 63.67 $\pm$ 4.26
& 35.83 $\pm$ 17.27
& 44.70
& 59.53
& 52.11 \\

& Base
& 56.11 $\pm$ 5.24
& 31.65 $\pm$ 14.45
& 58.40 $\pm$ 4.69
& 38.11 $\pm$ 14.54
& \uline{64.25}
& 13.65
& 38.95 \\

& SSR
& \uline{69.57} $\pm$ 9.35
& \uline{44.44} $\pm$ 25.82
& \uline{67.77} $\pm$ 6.38
& 41.55 $\pm$ 20.52
& 47.57
& \textbf{66.19}
& \uline{56.88} \\

\midrule

\multirow{4}{*}{Llama3.1-8B}
& Patient-$\psi$
& 61.39 $\pm$ 4.20
& 21.06 $\pm$ 10.27
& 62.03 $\pm$ 4.44
& 26.74 $\pm$ 10.15
& \textbf{64.64}
& 35.78
& 50.21 \\

& UPSD$^4$
& 63.56 $\pm$ 5.41
& 31.55 $\pm$ 18.57
& 64.38 $\pm$ 4.74
& 35.17 $\pm$ 17.38
& 41.64
& 63.85
& 52.74 \\

& Base
& 61.52 $\pm$ 5.21
& 25.88 $\pm$ 14.18
& 62.42 $\pm$ 4.39
& 30.64 $\pm$ 13.66
& 62.42
& 43.89
& 53.16 \\

& SSR
& \textbf{69.86} $\pm$ 9.38
& \textbf{45.91} $\pm$ 25.37
& \textbf{67.82} $\pm$ 6.67
& \uline{41.81} $\pm$ 20.79
& 53.84
& \uline{64.47}
& \textbf{59.15} \\

\bottomrule
\end{tabular}
}
\vspace{-0.5cm}
\end{table*}

\paragraph{Utterance-Level Metrics} For utterance-level assessment, we employ both automatic metrics and custom-designed indicators. For all metrics that require categorical stigma labels on generated outputs, we reuse the same context-aware \texttt{gpt-5} judge described in Section~\ref{subsec:extract}. Given the held-out dialogue context and the generated patient response, the judge first decides whether the response expresses self-stigma and, when stigma is present, assigns one of the five stigma subtypes. Thus, Table~\ref{tab:main experiments}, Figure~\ref{fig:stigma-heatmap}, and Table~\ref{tab:stigma_type_f1} are based on a consistent prompted-LLM labeling protocol rather than keyword matching.
\begin{itemize}
    \item \textbf{Semantic Fidelity}: We compute BERTScore~\cite{Zhang2020BERTScore:} and Sentence-BERT~\cite{reimers-gurevych-2019-sentence} Cosine Similarity against reference responses as auxiliary reference-based metrics for semantic relevance. Because therapeutic dialogue is inherently one-to-many, we do not treat these measures as decisive indicators of simulation quality.
    \item \textbf{Contextual Stigma F1}: We introduce Contextual Stigma F1 (\CSF{}) to evaluate whether a model expresses self-stigma only when the conversational context warrants it. We treat the stigma-triggering test set as the positive class and the neutral test set as the negative class. Thus, true positives ($TP$) are stigma expressions generated in stigma contexts, false negatives ($FN$) are non-stigma responses in stigma contexts, false positives ($FP$) are stigma expressions generated in neutral contexts, and true negatives ($TN$) are non-stigma responses in neutral contexts. We compute class-specific F1 scores as:\vspace{-0.1cm}
    \begin{align}
        \mathrm{CSF1}_{+} &= \frac{2TP}{2TP + FP + FN}, \\
        \mathrm{CSF1}_{-} &= \frac{2TN}{2TN + FP + FN}, \\
        \mathrm{CSF1}_{\text{macro}} &= \frac{\mathrm{CSF1}_{+} + \mathrm{CSF1}_{-}}{2}.
    \end{align}\vspace{-0.1cm}
    The positive score $\mathrm{CSF1}_{+}$ measures sensitivity to stigma-triggering contexts, while the negative score $\mathrm{CSF1}_{-}$ measures restraint in neutral contexts. We report both class-specific scores and $\mathrm{CSF1}_{\text{macro}}$ as an overall contextual-stigma indicator because the macro score balances both capabilities and avoids rewarding models that simply overproduce stigma regardless of context.

\end{itemize}

\paragraph{Semantic and Stigma Behavioral Analysis} To assess the clinical authenticity of the generated stigma expressions \textit{i.e.} how closely they mirror the linguistic and behavioral patterns observed in real patients, we perform a fine-grained semantic and behavioral analysis. 
\begin{itemize}
    \item \textbf{LIWC Analysis} We use Linguistic Inquiry and Word Count (LIWC) dictionaries, English~\cite{pennebaker2015development} and Chinese~\cite{Zeng_Yang_Tu_Liu_Sun_2018} to analyze linguistic markers (e.g., \textit{negative emotion, self-reference, tentativeness}) indicative of internalized stigma.
    \item \textbf{Stigma Type Classification} We evaluate the precision of generated stigma behaviors by calculating F1-scores for five clinically-grounded types: Avoidance, Denial, Self-blame, Defensiveness, and Social Concern~\cite{Amsalem2023-dz, doi:10.1176/appi.ps.201200561, Oexle_Ajdacic-Gross_Kilian_Müller_Rodgers_Xu_Rössler_Rüsch_2017, Prizeman2024-xn, Ociskova2023stigma} (see Appendix Table~\ref{tab:stigma-types} for definitions). .The predicted subtype for each generated response is the subtype assigned by the same \texttt{gpt-5} judge.

\end{itemize}

\paragraph{Personnel-Level Scale} 
To gauge the model's ability to elicit clinically relevant self-stigma, we use established psychometric scales, namely \textit{Self-Stigma of Mental Illness Scale} (SSMIS)~\cite{Corrigan2012-gz,doi:10.1521/jscp.2006.25.8.875} and  \textit{Internalized Stigma of Mental Illness Scale} (ISMI)~\cite{Ritsher2003-ja,Boyd2014-gx} to score simulated patient assessments.

\section{Results and Discussion}
Our experimental results combine reference-based semantic metrics, behavior-level analyses, psychometric scales, and expert human evaluation to test both the fidelity and the contextual sensitivity of SSR-trained patient agents. The findings are summarized in Table \ref{tab:main experiments} and Figure \ref{fig:stigma-heatmap}.

\subsection{SSR Improves Fidelity and Context Sensitivity~(RQ 1\&2)}
Table~\ref{tab:main experiments} shows that SSR consistently improves response quality and context sensitivity, but the strongest evidence comes from the matched-backbone comparisons rather than from the reference-based metrics alone.

\textbf{Auxiliary semantic gains on matched backbones.} Because BERTScore and SentenceBERT compare outputs against a single reference, we interpret them only as supportive signals. Even under that interpretation, SSR improves both metrics over same-backbone prompt baselines: for example, Qwen3-8B (SSR) reaches 69.57/44.44 on the neutral test set and 67.77/41.55 on the stigma test set, compared with 62.62/33.61 and 64.79/41.14 for Patient-$\psi$ on the same backbone.

\textbf{Clearer context separation from \CSF{}.} \CSF{} reveals the main behavioral difference by evaluating both sensitivity to stigma-triggering contexts and restraint in neutral contexts. On Qwen3-8B, SSR improves the macro score over the base model and all matched-backbone prompt baselines, with especially large gains on the negative class. On Llama3.1-8B, SSR also achieves the strongest macro score among matched-backbone systems. This pattern shows why class-specific evaluation is necessary: base models can obtain high positive-class scores by overproducing stigma-like responses, but their negative-class scores drop when they fail to suppress such behavior in neutral contexts. SSR models show stronger contextual separation because they preserve higher negative-class \CSF{} while maintaining competitive positive-class performance.

\textbf{Fairness-controlled improvement.} Applying the original prompt baselines directly to Qwen3-8B and Llama3.1-8B isolates the effect of the SSR objective from model-family differences. Under this control, the SSR-fine-tuned models remain the strongest systems on both backbones, supporting the claim that the performance gains come from the internal-monologue training signal rather than from a stronger underlying model family.

\subsection{SSR Generates Clinically Authentic Stigma Behaviors~(RQ3)}
\label{subsec:stigma_distribution}
Beyond overall quality, our analysis indicates that SSR-trained models produce stigma expressions that are both distributionally and linguistically aligned with the patterns observed in source data.
\begin{figure}
    \centering
    \includegraphics[width=\linewidth]{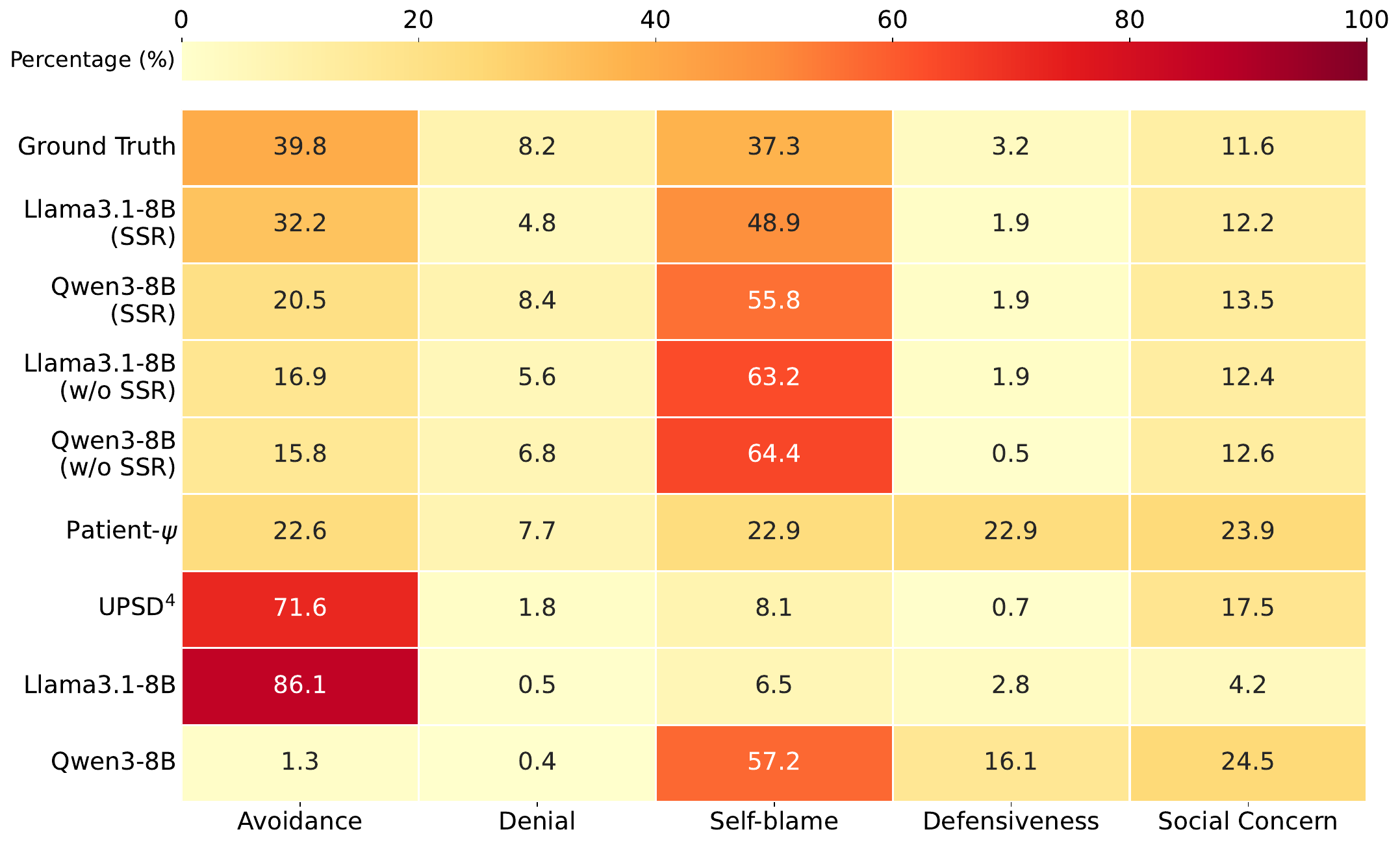}
    \vspace{-0.5cm}
    \caption{Distribution of stigma expression types across different models compared to ground truth annotations. Darker shades indicate higher proportion of each stigma type in model outputs.}
    \label{fig:stigma-heatmap}
\vspace{-0.5cm}
\end{figure}

\textbf{Clinically-Plausible Distribution of Stigma Types. }
Figure \ref{fig:stigma-heatmap} shows that SSR-trained models generate a balanced distribution of stigma types that aligns with ground truth clinical data. Crucially, they correct a critical flaw in baseline models—an over-reliance on Self-blame (reducing its proportion from ~64\% to ~50-55\%)—while appropriately increasing the prevalence of Avoidance and other nuanced types. This shift reflects a more accurate simulation of how self-stigma manifests in real patients.

\begin{table}[htbp]
\centering
\caption{F1-scores (\%) for five stigma‑type classifications. S.b.: Self-blame, S.C.: Social Concern}
\label{tab:stigma_type_f1}
\resizebox{\linewidth}{!}{
\begin{tabular}{lcccccc}
\toprule
\textbf{Models} & \textbf{Avoid} & \textbf{Deny} & \textbf{S.b.} & \textbf{Defend} & \textbf{S.C.} & \textbf{Avg.} \\
\midrule
Llama3.1-8B (SSR) & \textbf{49.7} & 25.8 & \textbf{68.0} & \textbf{25.0} & 30.8 & \textbf{39.9} \\
Qwen3-8B (SSR) & 41.7 & \textbf{41.2} & 67.9 & 0.0 & \textbf{36.4} & 37.4 \\
Llama3.1-8B (w/o SSR) & 24.0 & 0.0 & 46.1 & 0.0 & 13.3 & 16.7 \\
Qwen3-8B (w/o SSR) & 24.5 & 12.1 & 48.0 & 0.0 & 4.7 & 17.8 \\
Patient-$\psi$ & 25.4 & 12.5 & 24.2 & 5.3 & 9.3 & 15.3 \\
UPSD$^4$ & 52.5 & 0.0 & 18.9 & 0.0 & 14.5 & 17.2 \\
Llama3.1-8B & 56.0 & 0.0 & 20.8 & 0.0 & 11.4 & 17.6 \\
Qwen3-8B & 5.6 & 8.9 & 51.7 & 7.9 & 17.1 & 18.2 \\
\bottomrule
\end{tabular}
}
\end{table}

\textbf{Precision in Generating Specific Stigma Behaviors.} Table \ref{tab:stigma_type_f1} shows that SSR training improves the model's precision in generating specific, clinically-grounded stigma behaviors. Llama3.1-8B (SSR) achieves a macro-average F1-score of 39.9\%, a 21.7\% absolute improvement over the best non-SSR model. Notably, it shows marked gains on challenging behaviors like Denial (25.8\% F1) and Defensiveness (25.0\% F1), which most baselines fail to generate entirely (0\% F1), which indicates alignment makes defensive behavior hard; SSR improves but remains imperfect.


\subsection{Human Evaluation of Realism}
We further conduct an expert human evaluation to assess realism beyond automatic metrics. Three evaluators with graduate-level training in mental health or psychiatry and 1--5 years of relevant experience blindly rated 350 generated dialogues across neutral and stigma-triggering contexts on a 9-point scale for \textit{Authenticity}, \textit{Stigma Level}, \textit{Richness}, and \textit{Overall Quality} (full protocol and scores are reported in Table~\ref{tab:human-eval-full}). All evaluators were fairly compensated in accordance with the local minimum wage standards.

\begin{table}[!t]
\centering
\small
\caption{Full human evaluation results averaged across three expert raters.}
\vspace{-0.2cm}
\label{tab:human-eval-full}
\resizebox{\linewidth}{!}{
\begin{tabular}{lcccc}
\toprule
\textbf{Setting} & \textbf{Authenticity} & \textbf{Stigma Level} & \textbf{Richness} & \textbf{Overall Quality} \\
\midrule
Golden/Original & 5.74 & 3.39 & 4.03 & 6.03 \\
Golden/Stigma & 6.68 & 6.32 & 5.45 & 7.05 \\
Patient-$\psi$/Original & 5.58 & 4.58 & 5.91 & 5.95 \\
Patient-$\psi$/Stigma & 6.40 & 5.22 & 6.64 & 6.67 \\
UPSD$^4$/Original & 5.45 & 3.26 & 4.71 & 5.87 \\
UPSD$^4$/Stigma & 6.44 & 4.20 & 5.88 & 6.07 \\
SSR/Original & 6.09 & 2.84 & 5.00 & 6.09 \\
SSR/Stigma & 6.71 & 4.17 & 5.40 & 6.14 \\
\bottomrule
\end{tabular}}
\vspace{-0.5cm}
\end{table}

The human ratings provide three complementary findings. First, SSR achieves the highest human-rated authenticity among model in both neutral and stigma contexts, scoring 6.09 and 6.71, respectively. Second, SSR shows the clearest neutral-to-stigma shift in perceived stigma level among the model baselines, increasing from 2.84 to 4.17, compared with 4.58 to 5.22 for Patient-$\psi$ and 3.26 to 4.20 for UPSD$^4$, providing direct support that SSR improves perceived realism and context-sensitive stigma expression beyond what reference-based metrics alone can show.

\subsection{Scale Test Results}
\vspace{-0.2cm}
\begin{figure}[htbp]
    \centering
    \includegraphics[width=\linewidth,  trim=0 0.7cm 0 0,  clip]{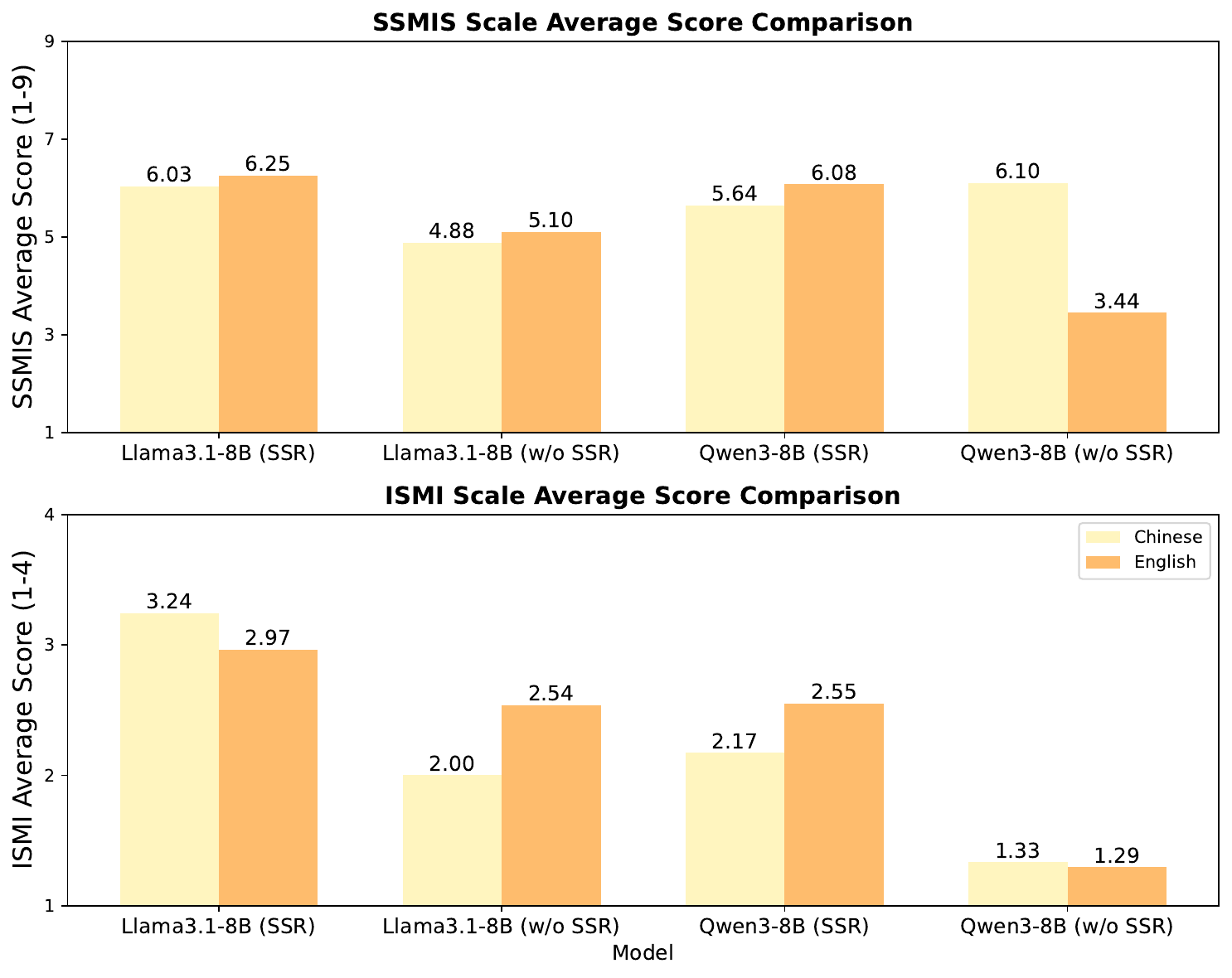}
    \vspace{-0.5cm}
    \caption{SSMIS and ISMI scale scores for SSR‑fine‑tuned and baseline models. SSR models produce higher stigma scores, reflecting more authentic self‑stigmatization in simulated patient responses.}
    \label{fig:scale}
\vspace{-0.3cm}
\end{figure}

The psychometric scale evaluations provide another view of whether the simulated persona occupies a meaningful self-stigma range rather than merely producing more resistant dialogue acts.

As illustrated in Figure \ref{fig:scale}, models fine-tuned with SSR achieve higher total and average scores on both the SSMIS and ISMI scales while maintaining near-perfect response validity. For instance, \textbf{Llama3.1-8B (SSR)} attains an SSMIS average score of 6.25 (EN) and 6.03 (CN), substantially surpassing its non-SSR counterpart (4.88 and 5.10, respectively). Models without SSR reasoning component exhibit lower scale scores and, critically, suffer from inconsistent response validity---as notably observed with Qwen3-8B (w/o SSR).

Using commonly applied ISMI interpretations~\cite{Lysaker2007-pr,Ritsher2004-mg}, scores from 1.00--2.00 indicate minimal or no internalized stigma, 2.01--2.50 mild stigma, 2.51--3.00 moderate stigma, and 3.01--4.00 severe stigma. Under this lens, models without SSR remain in a minimal range (e.g., Qwen3-8B w/o SSR at 1.29--1.33), whereas Qwen3-8B (SSR) shifts into a mild-to-moderate range (2.17--2.55) and Llama3.1-8B (SSR) reaches moderate-to-severe levels (2.97--3.24). This pattern suggests that SSR does not merely inflate scores; it moves the simulated persona into meaningful self-stigma regimes.

\subsection{Linguistic Style Analysis with LIWC}
\label{subsec:liwc}
To further investigate the linguistic shifts induced by SSR, we conduct a psycholinguistic analysis using LIWC~\cite{pennebaker2015development,Zeng_Yang_Tu_Liu_Sun_2018}. Across both English and Chinese, stigma-triggering responses become more self-focused and socially withdrawn: in English, negative emotion and first-person singular usage increase while social and assent words decrease; in Chinese, negation, exclamation, and cognitive-mechanism words increase while social terms decrease sharply. We treat LIWC as descriptive support rather than causal evidence, but the cross-lingual pattern is consistent with self-stigma as heightened self-focus and reduced social openness.

\section{Conclusion}
This work bridges a critical gap between computational dialogue modeling and the psychological reality of self-stigma by introducing the Stigmatized Self-Reflection (SSR) framework, which grounds patient simulation in the established 3A1H model of internalized stigma. By teaching models to generate internal stigma-aware reasoning before producing external responses, our approach yields more authentic and context-sensitive patient simulations than strong prompt-based baselines, with updated evidence that now combines automated analyses and expert human evaluation. We view this as a theory-informed step toward more realistic training-oriented patient agents rather than a deployment-ready clinical tool.

\newpage

\section*{Limitations}

While promising, our work has limitations. The reliance on GPT-5 for stigma annotation, though pragmatic, introduces potential biases. Our evaluation now combines automated analyses with expert human evaluation and binary stigma-label validation, but it still lacks clinician-only assessment and a full human audit of the multi-label annotation pipeline. Furthermore, our current framework simulates a generalized self-stigma process; future work could personalize simulations by modeling individual stigma trajectories based on patient history.

\section*{Ethical Considerations}

We recognize the sensitive nature of simulating mental health conditions. This tool is intended strictly for training and research under expert supervision, not for diagnosis or patient interaction. By creating more realistic and nuanced simulations, we aim to contribute to the education of empathetic clinicians and the development of support systems that better understand and address the profound impact of self-stigma. The mental health dialogue corpora used in this study (e.g., Alexander Street Transcripts, D$^4$) were obtained from publicly available sources or through established research data-sharing agreements. All data used have been thoroughly anonymized to remove any personally identifiable information (PII), in accordance with ethical guidelines for research involving human subjects. The construction of our Stigmatized Self-Reflection (SSR) dataset involved only synthetic augmentation of these anonymized utterances, ensuring no additional privacy risks were introduced.

While our work aims to advance mental health training, we acknowledge several potential risks that require careful consideration and mitigation. First, there is a risk of misuse or misinterpretation. If deployed without proper safeguards or expert oversight, the simulated patient could be mistaken for a diagnostic tool or used in contexts that replace essential human clinical judgment, potentially leading to harmful outcomes. Second, despite our efforts to enhance realism, the simulation remains an approximation. **Inaccuracies or biases** in the model’s portrayal of self-stigma could inadvertently reinforce stereotypes or provide trainees with misleading behavioral patterns. Third, the training data, though anonymized, may reflect societal biases present in the source dialogues. Without ongoing evaluation, the model could perpetuate these biases in its responses. Finally, the very capability to simulate vulnerable states raises concerns about potential exploitation, such as using the technology to manipulate or generate distressing content. We strongly advocate for the development and adherence to clear use-case guidelines, continuous validation of the model’s outputs by clinical experts, and the implementation of access controls to ensure this technology is used solely for its intended educational and research purposes.


\newpage
\bibliography{extracted}

@book{world2025mental,
  author =        {World Health Organization},
  publisher =     {World Health Organization},
  title =         {Mental health atlas 2024},
  year =          {2025},
}

@misc{WHO2025,
  author =        {World Health Organization},
  note =          {License: CC BY-NC-SA 3.0 IGO},
  publisher =     {World Health Organization},
  title =         {World mental health today: latest data},
  year =          {2025},
  url =           {https://iris.who.int/handle/10665/382343},
}

@inproceedings{yao-etal-2022-d4,
  address =       {Abu Dhabi, United Arab Emirates},
  author =        {Yao, Binwei and Shi, Chao and Zou, Likai and
                   Dai, Lingfeng and Wu, Mengyue and Chen, Lu and
                   Wang, Zhen and Yu, Kai},
  booktitle =     {Proceedings of the 2022 Conference on Empirical
                   Methods in Natural Language Processing},
  editor =        {Goldberg, Yoav and Kozareva, Zornitsa and Zhang, Yue},
  month =         dec,
  pages =         {2438--2459},
  publisher =     {Association for Computational Linguistics},
  title =         {D4: a {C}hinese Dialogue Dataset for
                   Depression-Diagnosis-Oriented Chat},
  year =          {2022},
  doi =           {10.18653/v1/2022.emnlp-main.156},
  url =           {https://aclanthology.org/2022.emnlp-main.156/},
}

@misc{chen2023llmempoweredchatbotspsychiatristpatient,
  author =        {Siyuan Chen and Mengyue Wu and Kenny Q. Zhu and
                   Kunyao Lan and Zhiling Zhang and Lyuchun Cui},
  title =         {LLM-empowered Chatbots for Psychiatrist and Patient
                   Simulation: Application and Evaluation},
  year =          {2023},
  url =           {https://arxiv.org/abs/2305.13614},
}

@misc{lan2024reliableempatheticdepressiondiagnosisorientedchats,
  author =        {Kunyao Lan and Cong Ming and Binwei Yao and Lu Chen and
                   Mengyue Wu},
  title =         {Towards Reliable and Empathetic
                   Depression-Diagnosis-Oriented Chats},
  year =          {2024},
  url =           {https://arxiv.org/abs/2404.05012},
}

@misc{lan2024depressiondiagnosisdialoguesimulation,
  author =        {Kunyao Lan and Bingrui Jin and Zichen Zhu and
                   Siyuan Chen and Shu Zhang and Kenny Q. Zhu and
                   Mengyue Wu},
  title =         {Depression Diagnosis Dialogue Simulation:
                   Self-improving Psychiatrist with Tertiary Memory},
  year =          {2024},
  url =           {https://arxiv.org/abs/2409.15084},
}

@inproceedings{sharma2024towards,
  author =        {Mrinank Sharma and Meg Tong and Tomasz Korbak and
                   David Duvenaud and Amanda Askell and Samuel R. Bowman and
                   Esin DURMUS and Zac Hatfield-Dodds and
                   Scott R Johnston and Shauna M Kravec and
                   Timothy Maxwell and Sam McCandlish and Kamal Ndousse and
                   Oliver Rausch and Nicholas Schiefer and Da Yan and
                   Miranda Zhang and Ethan Perez},
  booktitle =     {The Twelfth International Conference on Learning
                   Representations},
  title =         {Towards Understanding Sycophancy in Language Models},
  year =          {2024},
  url =           {https://openreview.net/forum?id=tvhaxkMKAn},
}

@article{Corrigan2002-se,
  author =        {Corrigan, Patrick W and Watson, Amy C},
  journal =       {World Psychiatry},
  month =         feb,
  number =        {1},
  pages =         {16--20},
  title =         {Understanding the impact of stigma on people with
                   mental illness},
  volume =        {1},
  year =          {2002},
  language =      {en},
}

@article{https://doi.org/10.1111/jasp.12323,
  author =        {Sadler, Melody S. and Kaye, Kimberly E. and
                   Vaughn, Allison A.},
  journal =       {Journal of Applied Social Psychology},
  number =        {11},
  pages =         {602-612},
  title =         {Competence and warmth stereotypes prompt mental
                   illness stigma through emotions},
  volume =        {45},
  year =          {2015},
  doi =           {https://doi.org/10.1111/jasp.12323},
  url =           {https://onlinelibrary.wiley.com/doi/abs/10.1111/jasp.12323},
}

@inproceedings{cao-etal-2025-breaking,
  address =       {Albuquerque, New Mexico},
  author =        {Cao, Jieming and Huang, Chen and Zhang, Yanan and
                   Deng, Ruibo and Zhang, Jincheng and Lei, Wenqiang},
  booktitle =     {Findings of the Association for Computational
                   Linguistics: NAACL 2025},
  editor =        {Chiruzzo, Luis and Ritter, Alan and Wang, Lu},
  month =         apr,
  pages =         {182--200},
  publisher =     {Association for Computational Linguistics},
  title =         {Breaking the Stigma! Unobtrusively Probe Symptoms in
                   Depression Disorder Diagnosis Dialogue},
  year =          {2025},
  doi =           {10.18653/v1/2025.findings-naacl.10},
  isbn =          {979-8-89176-195-7},
  url =           {https://aclanthology.org/2025.findings-naacl.10/},
}

@inproceedings{meng-etal-2025-stigma,
  address =       {Vienna, Austria},
  author =        {Meng, Han and Chen, Yancan and Li, Yunan and
                   Yang, Yitian and Lee, Jungup and Zhang, Renwen and
                   Lee, Yi-Chieh},
  booktitle =     {Proceedings of the 63rd Annual Meeting of the
                   Association for Computational Linguistics (Volume 1:
                   Long Papers)},
  editor =        {Che, Wanxiang and Nabende, Joyce and
                   Shutova, Ekaterina and Pilehvar, Mohammad Taher},
  month =         jul,
  pages =         {5453--5490},
  publisher =     {Association for Computational Linguistics},
  title =         {What is Stigma Attributed to? A Theory-Grounded,
                   Expert-Annotated Interview Corpus for Demystifying
                   Mental-Health Stigma},
  year =          {2025},
  doi =           {10.18653/v1/2025.acl-long.272},
  isbn =          {979-8-89176-251-0},
  url =           {https://aclanthology.org/2025.acl-long.272/},
}

@book{goffman1963stigma,
  author =        {Goffman, E.},
  publisher =     {Touchstone},
  title =         {Stigma: Notes on the Management of Spoiled Identity},
  year =          {1963},
  isbn =          {9780671622442},
  url =           {https://books.google.de/books?id=7CNUUMKTbIoC},
}

@inproceedings{wang-etal-2024-patient,
  address =       {Miami, Florida, USA},
  author =        {Wang, Ruiyi and Milani, Stephanie and Chiu, Jamie C. and
                   Zhi, Jiayin and Eack, Shaun M. and Labrum, Travis and
                   Murphy, Samuel M and Jones, Nev and Hardy, Kate V and
                   Shen, Hong and Fang, Fei and Chen, Zhiyu},
  booktitle =     {Proceedings of the 2024 Conference on Empirical
                   Methods in Natural Language Processing},
  editor =        {Al-Onaizan, Yaser and Bansal, Mohit and
                   Chen, Yun-Nung},
  month =         nov,
  pages =         {12772--12797},
  publisher =     {Association for Computational Linguistics},
  title =         {{PATIENT}-$\psi$: Using Large Language Models to
                   Simulate Patients for Training Mental Health
                   Professionals},
  year =          {2024},
  doi =           {10.18653/v1/2024.emnlp-main.711},
  url =           {https://aclanthology.org/2024.emnlp-main.711/},
}

@misc{ozgun2025trustworthyaipsychotherapymultiagent,
  author =        {Mithat Can Ozgun and Jiahuan Pei and Koen Hindriks and
                   Lucia Donatelli and Qingzhi Liu and Junxiao Wang},
  title =         {Trustworthy AI Psychotherapy: Multi-Agent LLM
                   Workflow for Counseling and Explainable Mental
                   Disorder Diagnosis},
  year =          {2025},
  url =           {https://arxiv.org/abs/2508.11398},
}

@article{10.1108/09504120910945317,
  author =        {Every‐Wurtz, Cheryl R.},
  journal =       {Reference Reviews},
  month =         {03},
  number =        {3},
  pages =         {39-40},
  title =         {Counseling and Psychotherapy Transcripts, Client
                   Narratives, and Reference Works},
  volume =        {23},
  year =          {2009},
  doi =           {10.1108/09504120910945317},
  issn =          {0950-4125},
  url =           {https://doi.org/10.1108/09504120910945317},
}

@inproceedings{li-etal-2023-understanding,
  address =       {Toronto, Canada},
  author =        {Li, Anqi and Ma, Lizhi and Mei, Yaling and
                   He, Hongliang and Zhang, Shuai and Qiu, Huachuan and
                   Lan, Zhenzhong},
  booktitle =     {Proceedings of the 61st Annual Meeting of the
                   Association for Computational Linguistics (Volume 1:
                   Long Papers)},
  editor =        {Rogers, Anna and Boyd-Graber, Jordan and
                   Okazaki, Naoaki},
  month =         jul,
  pages =         {10358--10376},
  publisher =     {Association for Computational Linguistics},
  title =         {Understanding Client Reactions in Online Mental
                   Health Counseling},
  year =          {2023},
  doi =           {10.18653/v1/2023.acl-long.577},
  url =           {https://aclanthology.org/2023.acl-long.577/},
}

@inproceedings{liu-etal-2021-towards,
  address =       {Online},
  author =        {Liu, Siyang and Zheng, Chujie and Demasi, Orianna and
                   Sabour, Sahand and Li, Yu and Yu, Zhou and
                   Jiang, Yong and Huang, Minlie},
  booktitle =     {Proceedings of the 59th Annual Meeting of the
                   Association for Computational Linguistics and the
                   11th International Joint Conference on Natural
                   Language Processing (Volume 1: Long Papers)},
  editor =        {Zong, Chengqing and Xia, Fei and Li, Wenjie and
                   Navigli, Roberto},
  month =         aug,
  pages =         {3469--3483},
  publisher =     {Association for Computational Linguistics},
  title =         {Towards Emotional Support Dialog Systems},
  year =          {2021},
  doi =           {10.18653/v1/2021.acl-long.269},
  url =           {https://aclanthology.org/2021.acl-long.269/},
}

@article{Amsalem2023-dz,
  author =        {Amsalem, Doron and Rogers, R Tyler and
                   Stroup, T Scott and Dixon, Lisa and Pope, Leah G},
  journal =       {Psychiatr. Rehabil. J.},
  month =         sep,
  number =        {3},
  pages =         {243--249},
  publisher =     {American Psychological Association (APA)},
  title =         {Self-stigma among people with serious mental
                   illnesses: The use of focus groups to inform the
                   development of a brief video intervention},
  volume =        {46},
  year =          {2023},
  language =      {en},
}

@article{doi:10.1176/appi.ps.201200561,
  author =        {Elise Pattyn and Mieke Verhaeghe and Charlotte Sercu and
                   Piet Bracke},
  journal =       {Psychiatric Services},
  number =        {2},
  pages =         {232-238},
  title =         {Public Stigma and Self-Stigma: Differential
                   Association With Attitudes Toward Formal and Informal
                   Help Seeking},
  volume =        {65},
  year =          {2014},
  doi =           {10.1176/appi.ps.201200561},
  url =           {https://psychiatryonline.org/doi/abs/10.1176/
                  appi.ps.201200561},
}

@article{Oexle_Ajdacic-Gross_Kilian_Müller_Rodgers_Xu_Rössler_Rüsch_2017,
  author =        {Oexle, N. and Ajdacic-Gross, V. and Kilian, R. and
                   Müller, M. and Rodgers, S. and Xu, Z. and
                   Rössler, W. and Rüsch, N.},
  journal =       {Epidemiology and Psychiatric Sciences},
  number =        {1},
  pages =         {53–60},
  title =         {Mental illness stigma, secrecy and suicidal ideation},
  volume =        {26},
  year =          {2017},
  doi =           {10.1017/S2045796015001018},
}

@article{Prizeman2024-xn,
  author =        {Prizeman, Katie and McCabe, Ciara and
                   Weinstein, Netta},
  journal =       {PLoS One},
  month =         jan,
  number =        {1},
  pages =         {e0296221},
  publisher =     {Public Library of Science (PLoS)},
  title =         {Stigma and its impact on disclosure and mental health
                   secrecy in young people with clinical depression
                   symptoms: A qualitative analysis},
  volume =        {19},
  year =          {2024},
  language =      {en},
}

@article{Ociskova2023stigma,
  author =        {Ociskova, Marie and Prasko, Jan and
                   Holubova, Michaela and Latalova, Klara and
                   Sollár, Tomáš and Zatkova, Marta and
                   Slepecky, Milos and Bocek, Jonas},
  journal =       {Neuro endocrinology letters},
  month =         {09},
  pages =         {368-383},
  title =         {Self-stigma in patients with schizophrenia: Impact
                   and management},
  volume =        {44},
  year =          {2023},
}

@inproceedings{chen-etal-2024-mapping,
  address =       {Mexico City, Mexico},
  author =        {Chen, Siyuan and Wang, Meilin and Lv, Minghao and
                   Zhang, Zhiling and Ju, Qianqian and Dejiyangla and
                   Peng, Yujia and Zhu, Kenny Q. and Wu, Mengyue},
  booktitle =     {Proceedings of the 2024 Conference of the North
                   American Chapter of the Association for Computational
                   Linguistics: Human Language Technologies (Volume 1:
                   Long Papers)},
  editor =        {Duh, Kevin and Gomez, Helena and Bethard, Steven},
  month =         jun,
  pages =         {5472--5487},
  publisher =     {Association for Computational Linguistics},
  title =         {Mapping Long-term Causalities in Psychiatric
                   Symptomatology and Life Events from Social Media},
  year =          {2024},
  doi =           {10.18653/v1/2024.naacl-long.306},
  url =           {https://aclanthology.org/2024.naacl-long.306/},
}

@article{10.1093/occmed/kqx099,
  author =        {Noone, Peter A},
  journal =       {Occupational Medicine},
  month =         {10},
  number =        {7},
  pages =         {581-582},
  title =         {The Holmes–Rahe Stress Inventory},
  volume =        {67},
  year =          {2017},
  doi =           {10.1093/occmed/kqx099},
  issn =          {0962-7480},
  url =           {https://doi.org/10.1093/occmed/kqx099},
}

@article{https://doi.org/10.1002/da.23155,
  author =        {Ruengorn, Chidchanok and Awiphan, Ratanaporn and
                   Wongpakaran, Nahathai and Wongpakaran, Tinakon and
                   Nochaiwong, Surapon and { for the Health Outcomes and
                     Mental Health Care Evaluation Survey Research Group
  (HOME-Survey) }},
  journal =       {Depression and Anxiety},
  number =        {6},
  pages =         {648-660},
  title =         {Association of job loss, income loss, and financial
                   burden with adverse mental health outcomes during
                   coronavirus disease 2019 pandemic in Thailand: A
                   nationwide cross-sectional study},
  volume =        {38},
  year =          {2021},
  doi =           {https://doi.org/10.1002/da.23155},
  url =           {https://onlinelibrary.wiley.com/doi/abs/10.1002/da.23155},
}

@inproceedings{zhang-etal-2022-symptom,
  address =       {Abu Dhabi, United Arab Emirates},
  author =        {Zhang, Zhiling and Chen, Siyuan and Wu, Mengyue and
                   Zhu, Kenny},
  booktitle =     {Proceedings of the 2022 Conference on Empirical
                   Methods in Natural Language Processing},
  editor =        {Goldberg, Yoav and Kozareva, Zornitsa and Zhang, Yue},
  month =         dec,
  pages =         {9970--9985},
  publisher =     {Association for Computational Linguistics},
  title =         {Symptom Identification for Interpretable Detection of
                   Multiple Mental Disorders on Social Media},
  year =          {2022},
  doi =           {10.18653/v1/2022.emnlp-main.677},
  url =           {https://aclanthology.org/2022.emnlp-main.677/},
}

@misc{grattafiori2024llama3herdmodels,
  author =        {Aaron Grattafiori and Abhimanyu Dubey and
                   Abhinav Jauhri and Abhinav Pandey and Abhishek Kadian and
                   Ahmad Al-Dahle and Aiesha Letman and Akhil Mathur and
                   Alan Schelten and Alex Vaughan and Amy Yang and
                   Angela Fan and Anirudh Goyal and Anthony Hartshorn and
                   Aobo Yang and Archi Mitra and Archie Sravankumar and
                   Artem Korenev and Arthur Hinsvark and Arun Rao and
                   Aston Zhang and Aurelien Rodriguez and
                   Austen Gregerson and Ava Spataru and Baptiste Roziere and
                   Bethany Biron and Binh Tang and Bobbie Chern and
                   Charlotte Caucheteux and Chaya Nayak and Chloe Bi and
                   Chris Marra and Chris McConnell and Christian Keller and
                   Christophe Touret and Chunyang Wu and Corinne Wong and
                   Cristian Canton Ferrer and Cyrus Nikolaidis and
                   Damien Allonsius and Daniel Song and Danielle Pintz and
                   Danny Livshits and Danny Wyatt and David Esiobu and
                   Dhruv Choudhary and Dhruv Mahajan and
                   Diego Garcia-Olano and Diego Perino and
                   Dieuwke Hupkes and Egor Lakomkin and Ehab AlBadawy and
                   Elina Lobanova and Emily Dinan and Eric Michael Smith and
                   Filip Radenovic and Francisco Guzmán and Frank Zhang and
                   Gabriel Synnaeve and Gabrielle Lee and
                   Georgia Lewis Anderson and Govind Thattai and
                   Graeme Nail and Gregoire Mialon and Guan Pang and
                   Guillem Cucurell and Hailey Nguyen and
                   Hannah Korevaar and Hu Xu and Hugo Touvron and
                   Iliyan Zarov and Imanol Arrieta Ibarra and
                   Isabel Kloumann and Ishan Misra and Ivan Evtimov and
                   Jack Zhang and Jade Copet and Jaewon Lee and
                   Jan Geffert and Jana Vranes and Jason Park and
                   Jay Mahadeokar and Jeet Shah and Jelmer van der Linde and
                   Jennifer Billock and Jenny Hong and Jenya Lee and
                   Jeremy Fu and Jianfeng Chi and Jianyu Huang and
                   Jiawen Liu and Jie Wang and Jiecao Yu and
                   Joanna Bitton and Joe Spisak and Jongsoo Park and
                   Joseph Rocca and Joshua Johnstun and Joshua Saxe and
                   Junteng Jia and Kalyan Vasuden Alwala and
                   Karthik Prasad and Kartikeya Upasani and Kate Plawiak and
                   Ke Li and Kenneth Heafield and Kevin Stone and
                   Khalid El-Arini and Krithika Iyer and Kshitiz Malik and
                   Kuenley Chiu and Kunal Bhalla and Kushal Lakhotia and
                   Lauren Rantala-Yeary and Laurens van der Maaten and
                   Lawrence Chen and Liang Tan and Liz Jenkins and
                   Louis Martin and Lovish Madaan and Lubo Malo and
                   Lukas Blecher and Lukas Landzaat and Luke de Oliveira and
                   Madeline Muzzi and Mahesh Pasupuleti and Mannat Singh and
                   Manohar Paluri and Marcin Kardas and
                   Maria Tsimpoukelli and Mathew Oldham and Mathieu Rita and
                   Maya Pavlova and Melanie Kambadur and Mike Lewis and
                   Min Si and Mitesh Kumar Singh and Mona Hassan and
                   Naman Goyal and Narjes Torabi and Nikolay Bashlykov and
                   Nikolay Bogoychev and Niladri Chatterji and
                   Ning Zhang and Olivier Duchenne and Onur Çelebi and
                   Patrick Alrassy and Pengchuan Zhang and Pengwei Li and
                   Petar Vasic and Peter Weng and Prajjwal Bhargava and
                   Pratik Dubal and Praveen Krishnan and
                   Punit Singh Koura and Puxin Xu and Qing He and
                   Qingxiao Dong and Ragavan Srinivasan and
                   Raj Ganapathy and Ramon Calderer and
                   Ricardo Silveira Cabral and Robert Stojnic and
                   Roberta Raileanu and Rohan Maheswari and
                   Rohit Girdhar and Rohit Patel and Romain Sauvestre and
                   Ronnie Polidoro and Roshan Sumbaly and Ross Taylor and
                   Ruan Silva and Rui Hou and Rui Wang and
                   Saghar Hosseini and Sahana Chennabasappa and
                   Sanjay Singh and Sean Bell and Seohyun Sonia Kim and
                   Sergey Edunov and Shaoliang Nie and Sharan Narang and
                   Sharath Raparthy and Sheng Shen and Shengye Wan and
                   Shruti Bhosale and Shun Zhang and Simon Vandenhende and
                   Soumya Batra and Spencer Whitman and Sten Sootla and
                   Stephane Collot and Suchin Gururangan and
                   Sydney Borodinsky and Tamar Herman and Tara Fowler and
                   Tarek Sheasha and Thomas Georgiou and Thomas Scialom and
                   Tobias Speckbacher and Todor Mihaylov and Tong Xiao and
                   Ujjwal Karn and Vedanuj Goswami and Vibhor Gupta and
                   Vignesh Ramanathan and Viktor Kerkez and
                   Vincent Gonguet and Virginie Do and Vish Vogeti and
                   Vítor Albiero and Vladan Petrovic and Weiwei Chu and
                   Wenhan Xiong and Wenyin Fu and Whitney Meers and
                   Xavier Martinet and Xiaodong Wang and Xiaofang Wang and
                   Xiaoqing Ellen Tan and Xide Xia and Xinfeng Xie and
                   Xuchao Jia and Xuewei Wang and Yaelle Goldschlag and
                   Yashesh Gaur and Yasmine Babaei and Yi Wen and
                   Yiwen Song and Yuchen Zhang and Yue Li and Yuning Mao and
                   Zacharie Delpierre Coudert and Zheng Yan and
                   Zhengxing Chen and Zoe Papakipos and Aaditya Singh and
                   Aayushi Srivastava and Abha Jain and Adam Kelsey and
                   Adam Shajnfeld and Adithya Gangidi and
                   Adolfo Victoria and Ahuva Goldstand and Ajay Menon and
                   Ajay Sharma and Alex Boesenberg and Alexei Baevski and
                   Allie Feinstein and Amanda Kallet and Amit Sangani and
                   Amos Teo and Anam Yunus and Andrei Lupu and
                   Andres Alvarado and Andrew Caples and Andrew Gu and
                   Andrew Ho and Andrew Poulton and Andrew Ryan and
                   Ankit Ramchandani and Annie Dong and Annie Franco and
                   Anuj Goyal and Aparajita Saraf and
                   Arkabandhu Chowdhury and Ashley Gabriel and
                   Ashwin Bharambe and Assaf Eisenman and Azadeh Yazdan and
                   Beau James and Ben Maurer and Benjamin Leonhardi and
                   Bernie Huang and Beth Loyd and Beto De Paola and
                   Bhargavi Paranjape and Bing Liu and Bo Wu and Boyu Ni and
                   Braden Hancock and Bram Wasti and Brandon Spence and
                   Brani Stojkovic and Brian Gamido and Britt Montalvo and
                   Carl Parker and Carly Burton and Catalina Mejia and
                   Ce Liu and Changhan Wang and Changkyu Kim and
                   Chao Zhou and Chester Hu and Ching-Hsiang Chu and
                   Chris Cai and Chris Tindal and
                   Christoph Feichtenhofer and Cynthia Gao and
                   Damon Civin and Dana Beaty and Daniel Kreymer and
                   Daniel Li and David Adkins and David Xu and
                   Davide Testuggine and Delia David and Devi Parikh and
                   Diana Liskovich and Didem Foss and Dingkang Wang and
                   Duc Le and Dustin Holland and Edward Dowling and
                   Eissa Jamil and Elaine Montgomery and
                   Eleonora Presani and Emily Hahn and Emily Wood and
                   Eric-Tuan Le and Erik Brinkman and Esteban Arcaute and
                   Evan Dunbar and Evan Smothers and Fei Sun and
                   Felix Kreuk and Feng Tian and Filippos Kokkinos and
                   Firat Ozgenel and Francesco Caggioni and
                   Frank Kanayet and Frank Seide and
                   Gabriela Medina Florez and Gabriella Schwarz and
                   Gada Badeer and Georgia Swee and Gil Halpern and
                   Grant Herman and Grigory Sizov and Guangyi and Zhang and
                   Guna Lakshminarayanan and Hakan Inan and
                   Hamid Shojanazeri and Han Zou and Hannah Wang and
                   Hanwen Zha and Haroun Habeeb and Harrison Rudolph and
                   Helen Suk and Henry Aspegren and Hunter Goldman and
                   Hongyuan Zhan and Ibrahim Damlaj and Igor Molybog and
                   Igor Tufanov and Ilias Leontiadis and
                   Irina-Elena Veliche and Itai Gat and Jake Weissman and
                   James Geboski and James Kohli and Janice Lam and
                   Japhet Asher and Jean-Baptiste Gaya and Jeff Marcus and
                   Jeff Tang and Jennifer Chan and Jenny Zhen and
                   Jeremy Reizenstein and Jeremy Teboul and
                   Jessica Zhong and Jian Jin and Jingyi Yang and
                   Joe Cummings and Jon Carvill and Jon Shepard and
                   Jonathan McPhie and Jonathan Torres and Josh Ginsburg and
                   Junjie Wang and Kai Wu and Kam Hou U and Karan Saxena and
                   Kartikay Khandelwal and Katayoun Zand and
                   Kathy Matosich and Kaushik Veeraraghavan and
                   Kelly Michelena and Keqian Li and Kiran Jagadeesh and
                   Kun Huang and Kunal Chawla and Kyle Huang and
                   Lailin Chen and Lakshya Garg and Lavender A and
                   Leandro Silva and Lee Bell and Lei Zhang and
                   Liangpeng Guo and Licheng Yu and Liron Moshkovich and
                   Luca Wehrstedt and Madian Khabsa and Manav Avalani and
                   Manish Bhatt and Martynas Mankus and Matan Hasson and
                   Matthew Lennie and Matthias Reso and Maxim Groshev and
                   Maxim Naumov and Maya Lathi and Meghan Keneally and
                   Miao Liu and Michael L. Seltzer and Michal Valko and
                   Michelle Restrepo and Mihir Patel and Mik Vyatskov and
                   Mikayel Samvelyan and Mike Clark and Mike Macey and
                   Mike Wang and Miquel Jubert Hermoso and Mo Metanat and
                   Mohammad Rastegari and Munish Bansal and
                   Nandhini Santhanam and Natascha Parks and
                   Natasha White and Navyata Bawa and Nayan Singhal and
                   Nick Egebo and Nicolas Usunier and Nikhil Mehta and
                   Nikolay Pavlovich Laptev and Ning Dong and
                   Norman Cheng and Oleg Chernoguz and Olivia Hart and
                   Omkar Salpekar and Ozlem Kalinli and Parkin Kent and
                   Parth Parekh and Paul Saab and Pavan Balaji and
                   Pedro Rittner and Philip Bontrager and Pierre Roux and
                   Piotr Dollar and Polina Zvyagina and
                   Prashant Ratanchandani and Pritish Yuvraj and
                   Qian Liang and Rachad Alao and Rachel Rodriguez and
                   Rafi Ayub and Raghotham Murthy and Raghu Nayani and
                   Rahul Mitra and Rangaprabhu Parthasarathy and
                   Raymond Li and Rebekkah Hogan and Robin Battey and
                   Rocky Wang and Russ Howes and Ruty Rinott and
                   Sachin Mehta and Sachin Siby and Sai Jayesh Bondu and
                   Samyak Datta and Sara Chugh and Sara Hunt and
                   Sargun Dhillon and Sasha Sidorov and Satadru Pan and
                   Saurabh Mahajan and Saurabh Verma and Seiji Yamamoto and
                   Sharadh Ramaswamy and Shaun Lindsay and Shaun Lindsay and
                   Sheng Feng and Shenghao Lin and Shengxin Cindy Zha and
                   Shishir Patil and Shiva Shankar and Shuqiang Zhang and
                   Shuqiang Zhang and Sinong Wang and Sneha Agarwal and
                   Soji Sajuyigbe and Soumith Chintala and Stephanie Max and
                   Stephen Chen and Steve Kehoe and Steve Satterfield and
                   Sudarshan Govindaprasad and Sumit Gupta and
                   Summer Deng and Sungmin Cho and Sunny Virk and
                   Suraj Subramanian and Sy Choudhury and Sydney Goldman and
                   Tal Remez and Tamar Glaser and Tamara Best and
                   Thilo Koehler and Thomas Robinson and Tianhe Li and
                   Tianjun Zhang and Tim Matthews and Timothy Chou and
                   Tzook Shaked and Varun Vontimitta and Victoria Ajayi and
                   Victoria Montanez and Vijai Mohan and
                   Vinay Satish Kumar and Vishal Mangla and Vlad Ionescu and
                   Vlad Poenaru and Vlad Tiberiu Mihailescu and
                   Vladimir Ivanov and Wei Li and Wenchen Wang and
                   Wenwen Jiang and Wes Bouaziz and Will Constable and
                   Xiaocheng Tang and Xiaojian Wu and Xiaolan Wang and
                   Xilun Wu and Xinbo Gao and Yaniv Kleinman and
                   Yanjun Chen and Ye Hu and Ye Jia and Ye Qi and
                   Yenda Li and Yilin Zhang and Ying Zhang and Yossi Adi and
                   Youngjin Nam and Yu and Wang and Yu Zhao and
                   Yuchen Hao and Yundi Qian and Yunlu Li and Yuzi He and
                   Zach Rait and Zachary DeVito and Zef Rosnbrick and
                   Zhaoduo Wen and Zhenyu Yang and Zhiwei Zhao and
                   Zhiyu Ma},
  title =         {The Llama 3 Herd of Models},
  year =          {2024},
  url =           {https://arxiv.org/abs/2407.21783},
}

@misc{yang2025qwen3technicalreport,
  author =        {An Yang and Anfeng Li and Baosong Yang and
                   Beichen Zhang and Binyuan Hui and Bo Zheng and
                   Bowen Yu and Chang Gao and Chengen Huang and
                   Chenxu Lv and Chujie Zheng and Dayiheng Liu and
                   Fan Zhou and Fei Huang and Feng Hu and Hao Ge and
                   Haoran Wei and Huan Lin and Jialong Tang and
                   Jian Yang and Jianhong Tu and Jianwei Zhang and
                   Jianxin Yang and Jiaxi Yang and Jing Zhou and
                   Jingren Zhou and Junyang Lin and Kai Dang and
                   Keqin Bao and Kexin Yang and Le Yu and Lianghao Deng and
                   Mei Li and Mingfeng Xue and Mingze Li and Pei Zhang and
                   Peng Wang and Qin Zhu and Rui Men and Ruize Gao and
                   Shixuan Liu and Shuang Luo and Tianhao Li and
                   Tianyi Tang and Wenbiao Yin and Xingzhang Ren and
                   Xinyu Wang and Xinyu Zhang and Xuancheng Ren and
                   Yang Fan and Yang Su and Yichang Zhang and
                   Yinger Zhang and Yu Wan and Yuqiong Liu and
                   Zekun Wang and Zeyu Cui and Zhenru Zhang and
                   Zhipeng Zhou and Zihan Qiu},
  title =         {Qwen3 Technical Report},
  year =          {2025},
  url =           {https://arxiv.org/abs/2505.09388},
}

@inproceedings{Zhang2020BERTScore:,
  author =        {Tianyi Zhang and Varsha Kishore and Felix Wu and
                   Kilian Q. Weinberger and Yoav Artzi},
  booktitle =     {International Conference on Learning Representations},
  title =         {BERTScore: Evaluating Text Generation with BERT},
  year =          {2020},
  url =           {https://openreview.net/forum?id=SkeHuCVFDr},
}

@inproceedings{reimers-gurevych-2019-sentence,
  address =       {Hong Kong, China},
  author =        {Reimers, Nils and Gurevych, Iryna},
  booktitle =     {Proceedings of the 2019 Conference on Empirical
                   Methods in Natural Language Processing and the 9th
                   International Joint Conference on Natural Language
                   Processing (EMNLP-IJCNLP)},
  editor =        {Inui, Kentaro and Jiang, Jing and Ng, Vincent and
                   Wan, Xiaojun},
  month =         nov,
  pages =         {3982--3992},
  publisher =     {Association for Computational Linguistics},
  title =         {Sentence-{BERT}: Sentence Embeddings using {S}iamese
                   {BERT}-Networks},
  year =          {2019},
  doi =           {10.18653/v1/D19-1410},
  url =           {https://aclanthology.org/D19-1410/},
}

@article{pennebaker2015development,
  author =        {Pennebaker, James W and Boyd, Ryan L and
                   Jordan, Kayla and Blackburn, Kate},
  title =         {The development and psychometric properties of
                   LIWC2015},
  year =          {2015},
}

@article{Zeng_Yang_Tu_Liu_Sun_2018,
  author =        {Zeng, Xiangkai and Yang, Cheng and Tu, Cunchao and
                   Liu, Zhiyuan and Sun, Maosong},
  journal =       {Proceedings of the AAAI Conference on Artificial
                   Intelligence},
  month =         {Apr.},
  number =        {1},
  title =         {Chinese LIWC Lexicon Expansion via Hierarchical
                   Classification of Word Embeddings with Sememe
                   Attention},
  volume =        {32},
  year =          {2018},
  doi =           {10.1609/aaai.v32i1.11982},
  url =           {https://ojs.aaai.org/index.php/AAAI/article/view/11982},
}

@article{Corrigan2012-gz,
  author =        {Corrigan, Patrick W and Michaels, Patrick J and
                   Vega, Eduardo and Gause, Michael and Watson, Amy C and
                   R{\"u}sch, Nicolas},
  journal =       {Psychiatry Res.},
  month =         aug,
  number =        {1},
  pages =         {65--69},
  publisher =     {Elsevier BV},
  title =         {Self-stigma of mental illness scale--short form:
                   reliability and validity},
  volume =        {199},
  year =          {2012},
  language =      {en},
}

@article{doi:10.1521/jscp.2006.25.8.875,
  author =        {Corrigan, Patrick W. and Watson, Amy C. and
                   Barr, Leah},
  journal =       {Journal of Social and Clinical Psychology},
  number =        {8},
  pages =         {875-884},
  title =         {The Self–Stigma of Mental Illness: Implications for
                   Self–Esteem and Self–Efficacy},
  volume =        {25},
  year =          {2006},
  doi =           {10.1521/jscp.2006.25.8.875},
  url =           {https://doi.org/10.1521/jscp.2006.25.8.875},
}

@article{Ritsher2003-ja,
  author =        {Ritsher, Jennifer Boyd and Otilingam, Poorni G and
                   Grajales, Monica},
  journal =       {Psychiatry Res.},
  month =         nov,
  number =        {1},
  pages =         {31--49},
  publisher =     {Elsevier BV},
  title =         {Internalized stigma of mental illness: psychometric
                   properties of a new measure},
  volume =        {121},
  year =          {2003},
  language =      {en},
}

@article{Boyd2014-gx,
  author =        {Boyd, Jennifer E and Adler, Emerald P and
                   Otilingam, Poorni G and Peters, Townley},
  journal =       {Compr. Psychiatry},
  month =         jan,
  number =        {1},
  pages =         {221--231},
  publisher =     {Elsevier BV},
  title =         {Internalized Stigma of Mental Illness ({ISMI}) scale:
                   a multinational review},
  volume =        {55},
  year =          {2014},
  language =      {en},
}

@ARTICLE{Lysaker2007-pr,
  title     = "Toward understanding the insight paradox: internalized stigma
               moderates the association between insight and social
               functioning, hope, and self-esteem among people with
               schizophrenia spectrum disorders",
  author    = "Lysaker, Paul H and Roe, David and Yanos, Philip T",
  journal   = "Schizophr. Bull.",
  publisher = "Oxford University Press (OUP)",
  volume    =  33,
  number    =  1,
  pages     = "192--199",
  month     =  jan,
  year      =  2007,
  language  = "en"
}

@ARTICLE{Ritsher2004-mg,
  title     = "Internalized stigma predicts erosion of morale among psychiatric
               outpatients",
  author    = "Ritsher, Jennifer Boyd and Phelan, Jo C",
  journal   = "Psychiatry Res.",
  publisher = "Elsevier BV",
  volume    =  129,
  number    =  3,
  pages     = "257--265",
  month     =  dec,
  year      =  2004,
  language  = "en"
}

\newpage
\appendix

\section{Reproducibility and Validation}
\label{appen:repro}

\subsection{Reproducibility of Stigma Annotation and SSR Generation}
All GPT-based annotation and generation runs used to build the SSR training data were executed in December 2025. Contextual stigma annotation was performed with \texttt{gpt-5}, where each candidate patient turn was evaluated jointly with the preceding dialogue history rather than in isolation. Under this rule, even a single manifestation (e.g., brief avoidance or sudden self-blame) could be labeled as self-stigma when the surrounding context showed that the utterance internalized negative beliefs or anticipated devaluation. The same \texttt{gpt-5} annotation protocol was reused to label generated outputs during evaluation: each output was judged for binary stigma presence and, when present, assigned one of the five stigma subtypes used in the training annotations. For SSR construction, every stigma-labeled utterance was paired with an internal monologue generated by \texttt{gpt-5-chat-latest}. At the time of collection, the closest static snapshot reference available for this system family was \texttt{gpt-5.1-2025-11-13}. Unless otherwise noted, SSR generation used \texttt{temperature = 0}, with all other decoding parameters kept at their default values.

\paragraph{Dialogue-level splitting and sampling.}
Before annotation, augmentation, or evaluation-instance construction, every original conversation was assigned a unique \texttt{dialogue\_id}. Because stigma-labeled conversations are rarer than neutral conversations, we kept all stigma-labeled \texttt{dialogue\_id}s and sampled neutral \texttt{dialogue\_id}s from the same corpora to produce a balanced set whose source-corpus, language, life-event, and symptom distributions match the stigma subset as closely as possible. Train/test partitioning was performed over \texttt{dialogue\_id}s rather than individual turns, so all turns and augmented instances derived from one conversation are confined to a single split. This prevents the model from seeing one turn of a conversation during fine-tuning and being evaluated on another turn from the same dialogue.

\paragraph{SSR generation prompt.}
The prompt instantiates the following fields: \texttt{<3a1h\_model>}, \texttt{<conversation\_history>}, \texttt{<patient\_response>}, \texttt{<stigma\_type>}, \texttt{<resistance\_level>}, and \texttt{<stigma\_indicators>}. The generation instruction used during rebuttal-period data construction is reproduced below in condensed form:

\begin{quote}
\footnotesize
You are writing the patient's private internal monologue before they say the final response. Given the 3A1H profile, the prior conversation history, the target patient response, the stigma type, the resistance level, and the stigma indicators, produce a 200--400 word first-person inner narrative that explains how the patient arrives at the final reply. The monologue should remain an internal thought process rather than an external analysis. It should naturally reflect awareness, agreement, application, and harm without naming those stages explicitly. It must match the language of the dialogue, so Chinese dialogues should receive Chinese monologues and English dialogues should receive English monologues. The output should support the final patient response but must not copy the target response or the provided labels verbatim. Output only the internal monologue.
\end{quote}

\section{Mental Health Corpus}
\label{appen:corpus}
\paragraph{Alexander Street Transcripts}
The Alexander Street Counseling and Therapy Transcripts~\cite{10.1108/09504120910945317} series is a comprehensive dataset comprising real-world therapeutic sessions, offering a distinct contrast to role-playing data. Unlike datasets constructed through controlled simulation, Alexander Street consists of multi-turn dialogues between licensed clinicians and actual clients, capturing the raw, unstructured dynamics of mental health consultations. The dataset covers a broad spectrum of psychological conditions, providing high-fidelity linguistic patterns of patient behaviors. Crucially for this study, these authentic transcripts contain naturally occurring markers of self-stigma—such as hesitation, meaningful silence, and defensive phrasing—serving as a vital reference for evaluating the authenticity and human-like resistance of our simulated models.

\paragraph{D$^4$ Dataset}
D$^4$~\cite{yao-etal-2022-d4} is a Chinese depression diagnosis dialogue dataset that closely mirrors real-world clinical practice. It consists of 1,339 dialogues between well-trained psychiatrists and simulated patients, with an average of 21.6 turns and about 61 utterances per conversation. Each dialogue is annotated by professional psychiatrists with accurate labels indicating the patient’s depression severity, categorized into four levels: non-depression, mild, moderate, and severe. In addition to diagnostic annotations, subsequent work has further enriched D$^4$ by constructing user profiles that summarize patients’ symptom based on DSM-5~\cite{lan2024reliableempatheticdepressiondiagnosisorientedchats}.

Given these precise expert annotations and well-defined patient profiles, D$^4$ serves as an ideal foundation for building and evaluating simulated patient agents, particularly for generating realistic and condition-aligned initial personas in psychiatric role-playing settings.

\paragraph{Client Reaction Dataset}
The Client Reaction dataset~\cite{li-etal-2023-understanding} is a collection of real client reactions extracted from online mental health counseling conversations. It contains 2,382 therapist-client dialogue turns annotated with fine-grained client reaction labels, including categories such as defending, confirming, and expressing confusion. The dataset is designed to capture nuanced client behaviors and responses in therapeutic settings, providing a valuable resource for training and evaluating dialogue systems on realistic human reactions. Unlike simulated or role-play data, the Client Reaction dataset reflects authentic client behaviors in digital mental health interventions, enabling research on responsive counseling systems that can adapt to diverse client needs and communication styles.

\paragraph{Emotional Support Conversation Dataset}
The Emotional Support Conversation (ESConv) dataset~\cite{liu-etal-2021-towards} is a large-scale collection of multi-turn emotional support dialogues collected from online support platforms. It consists of over 1,053 conversations between help-seekers and emotional supporters. Each conversation is annotated with detailed labels including seeker's emotional state, supporter's response strategies, and conversation outcomes. The dataset covers a wide range of emotional distress scenarios, from everyday stressors to severe psychological challenges. ESConv provides comprehensive resources for developing and evaluating emotional support dialogue systems, particularly focusing on how to effectively provide empathy, validation, and practical support in digital conversational settings.

\begin{table*}[t]
\centering
\scriptsize
\caption{Standardized Symptom Labels with Key Indicators}
\label{tab:symptoms}
\rowcolors{2}{gray!10}{white}
\begin{tabular}{p{0.30\linewidth} p{0.64\linewidth}}
\toprule
\textbf{Symptom Label} & \textbf{Key Indicators} \\
\midrule
Anger Irritability &
Feeling irritable or angry; emotional upset when reminded of trauma; verbal or physical aggression \\

Anxious Mood &
Excessive worry; persistent anxiety about events or panic; fear of traumatic events \\

Autonomic Symptoms &
Dizziness or light-headedness; accelerated heart rate; sweating or dry mouth; shortness of breath \\

Cardiovascular Symptoms &
Palpitations; chest pain; throbbing vessels; fainting sensations \\

Catatonic Behavior &
Hypervigilance; exaggerated startle response; being jumpy; strong physiological reactions to reminders \\

Decreased Energy Tiredness Fatigue &
Easily fatigued; lack of energy; sustained tiredness; reduced task efficiency \\

Depressed Mood &
Sad or hopeless mood; tearfulness; emotional numbness; feelings of emptiness \\

Gastrointestinal Symptoms &
Difficulty swallowing; nausea or vomiting; abdominal pain; diarrhea or constipation \\

Genitourinary Symptoms &
Loss of libido; menstrual irregularity; impotence; amenorrhea \\

Hyperactivity Agitation &
Restlessness; fidgeting; inability to sit still; pacing behaviors \\

Impulsivity &
Blurting out answers; difficulty waiting turn; interrupting others \\

Inattention &
Poor concentration; distractibility; difficulty organizing tasks; diminished focus \\

Indecisiveness &
Difficulty making decisions \\

Respiratory Symptoms &
Chest tightness; choking feelings; shortness of breath; breathing discomfort \\

Suicidal Ideas &
Recurrent thoughts of death; suicidal ideation or planning \\

Worthlessness and Guilty &
Excessive guilt; negative self-evaluation; self-reproach; exaggerated responsibility \\

Avoidance of Stimuli &
Avoiding trauma-related memories, thoughts, or external reminders \\

Compensatory Behaviors to Prevent Weight Gain &
Self-induced vomiting; laxative misuse; excessive exercise; fasting \\

Compulsions &
Repetitive behaviors; checking, cleaning, ordering, or counting rituals \\

Diminished Emotional Expression &
Emotional numbness; restrained expression; limited emotional display \\

Do Things Easily Get Painful Consequences &
Reckless behaviors; substance misuse; risky driving or sexual behavior \\

Drastical Shift in Mood and Energy &
Sudden changes in mood and energy levels \\

Fear About Social Situations &
Avoidance of social situations; fear of conversations or performance \\

Fear of Gaining Weight &
Intense fear of weight gain; body checking; weight-based self-evaluation \\

Fears of Being Negatively Evaluated &
Concern about negative judgment; embarrassment; avoidance of evaluative situations \\

Flight of Ideas &
Racing thoughts; pressured speech; abrupt topic shifts \\

Intrusion Symptoms &
Flashbacks; intrusive memories; distressing dreams; trauma reminders \\

Loss of Interest or Motivation &
Reduced enjoyment; social withdrawal; diminished interest in activities \\

More Talkative &
Excessive or pressured speech; talking without regard for others \\

Obsession &
Persistent unwanted thoughts; contamination fears; food-related preoccupation \\

Panic Fear &
Fear of losing control; fear of dying \\

Pessimism &
Hopeless outlook; negative expectations about the future \\

Poor Memory &
Forgetfulness; difficulty recalling daily events; memory gaps \\

Sleep Disturbance &
Insomnia or hypersomnia; nightmares; restless or unsatisfying sleep \\

Somatic Muscle &
Muscle tension; trembling; soreness; teeth grinding \\

Somatic Symptoms Others &
Depersonalization; numbness; uncontrollable crying or screaming \\

Somatic Symptoms Sensory &
Chills or heat sensations; blurred vision; weakness \\

Weight and Appetite Change &
Significant weight change; appetite disturbance; binge eating or restriction \\
\bottomrule
\end{tabular}
\end{table*}

\begin{table*}[t]
\centering
\scriptsize
\caption{Life Event Categories and Subcategories}
\label{tab:life-events}
\rowcolors{2}{gray!10}{white}
\begin{tabular}{p{0.30\linewidth} p{0.64\linewidth}}
\toprule
\textbf{Life Event Category} & \textbf{Subcategories} \\
\midrule
Health &
Accident or illness; mental illness; hospitalization; pregnancy; abuse; self-harm; suicide attempt; substance abuse; recovery; medication \\

Financial &
Financial difficulty or gain; loans; bankruptcy; major purchases; foreclosure; property loss \\

Relocation &
Moving within or across cities, states, or countries; refugee relocation; homelessness; major travel \\

Legal &
Arrest; legal action; incarceration; release from prison; minor law violations \\

Relationships Changes &
Marriage; divorce; breakup; reconciliation; abusive relationship; family conflict; parenting difficulties \\

New Birth in Family &
Birth of a child; adoption; becoming a grandparent or aunt/uncle \\

Death &
Death of spouse, child, parent, pet, friend, or extended family member \\

Career &
New job; promotion or demotion; job loss; workplace conflict; retirement; entrepreneurship \\

Education &
School or college transition; examination; certification; denial of admission \\

Lifestyle Change &
Change in habits or responsibilities; living condition changes; new pet; military service; vacation \\

Identity &
Gender or sexual identity; coming out; belief changes; major identity shift \\

Societal &
Natural disaster; pandemic; war; major political events with personal impact \\
\bottomrule
\end{tabular}
\end{table*}

\section{Extra Tables}
\begin{table*}[htbp]
\centering
\small
\caption{Types of Stigma Expressions in Patient Discourse}
\label{tab:stigma-types}
\begin{tabular}{l p{0.42\linewidth} p{0.34\linewidth}}
\toprule
\textbf{Stigma Type} & \textbf{Description} & \textbf{Example} \\
\midrule
\textbf{Avoidance} & 
Patient explicitly expresses unwillingness to discuss certain topics or avoids sensitive issues. &
``Well... I'd rather not talk about that.'' ``Can we discuss something else?'' \\

\rowcolor{gray!10}
\textbf{Denial} & 
Patient denies the existence or severity of the problem, claiming to be fine. &
``I'm not sick at all.'' ``I'm just a little tired, that's all.'' \\

\textbf{Self-blame} & 
Patient expresses self-deprecation, guilt, or shame regarding their condition or situation. &
``It's all my fault.'' ``I'm completely useless.'' \\

\rowcolor{gray!10}
\textbf{Defensiveness} & 
Patient exhibits defensive or aggressive responses toward diagnoses or questioning. &
``Why do you keep asking about this?'' ``Are you suggesting there's something wrong with me?'' \\

\textbf{Social Concern} & 
Patient expresses worry about others' perceptions and fears of social exclusion or judgment. &
``What will people think of me?'' ``I don't want anyone to know about this.'' \\
\bottomrule
\end{tabular}
\end{table*}

\begin{table*}[htbp]
\centering
\small
\caption{Top 10 LIWC category changes in English (ENG) between Stigma and Neutral conditions. Positive values indicate increased usage in stigma contexts.}
\label{tab:liwc-eng}
\begin{tabular}{l r r r r}
\toprule
\textbf{Category} & \textbf{Original} & \textbf{Stigma} & \textbf{Difference} & \textbf{\% Change} \\
\midrule
\multicolumn{5}{c}{\textbf{Largest Increases}} \\
\midrule
 present & 0.1074 & 0.1202 & +0.0128 & +11.92\% \\
   ipron & 0.0700 & 0.0820 & +0.0120 & +17.15\% \\
   verbs & 0.1497 & 0.1574 & +0.0077 & +5.16\% \\
 insight & 0.0521 & 0.0584 & +0.0063 & +12.13\% \\
  affect & 0.0727 & 0.0788 & +0.0061 & +8.45\% \\
  negemo & 0.0240 & 0.0297 & +0.0057 & +23.74\% \\
       i & 0.0154 & 0.0205 & +0.0051 & +32.91\% \\
  achiev & 0.0133 & 0.0183 & +0.0050 & +37.79\% \\
 percept & 0.0323 & 0.0366 & +0.0043 & +13.39\% \\
    hear & 0.0080 & 0.0107 & +0.0027 & +33.88\% \\
\midrule
\multicolumn{5}{c}{\textbf{Largest Decreases}} \\
\midrule
  social & 0.0902 & 0.0651 & -0.0251 & -27.82\% \\
  assent & 0.0310 & 0.0109 & -0.0201 & -64.80\% \\
   funct & 0.4595 & 0.4425 & -0.0170 & -3.69\% \\
   ppron & 0.0581 & 0.0469 & -0.0112 & -19.36\% \\
     you & 0.0257 & 0.0161 & -0.0095 & -37.16\% \\
 adverbs & 0.0738 & 0.0652 & -0.0086 & -11.69\% \\
  nonflu & 0.0108 & 0.0029 & -0.0079 & -73.12\% \\
    conj & 0.0599 & 0.0521 & -0.0078 & -13.06\% \\
   shehe & 0.0142 & 0.0081 & -0.0061 & -43.01\% \\
    prep & 0.1002 & 0.0953 & -0.0049 & -4.87\% \\
\bottomrule
\end{tabular}
\end{table*}

\begin{table*}[htbp]
\centering
\small
\caption{Top 10 LIWC category changes in Chinese (CHI) between Stigma and Neutral conditions.}
\label{tab:liwc-chi}
\begin{tabular}{l r r r r}
\toprule
\textbf{Category} & \textbf{Original} & \textbf{Stigma} & \textbf{Difference} & \textbf{\% Change} \\
\midrule
\multicolumn{5}{c}{\textbf{Largest Increases}} \\
\midrule
   funct & 0.0131 & 0.0222 & +0.0091 & +69.14\% \\
  negate & 0.0114 & 0.0198 & +0.0084 & +73.25\% \\
    excl & 0.0122 & 0.0206 & +0.0083 & +68.15\% \\
 cogmech & 0.0193 & 0.0273 & +0.0080 & +41.47\% \\
  negemo & 0.0034 & 0.0045 & +0.0011 & +32.46\% \\
  affect & 0.0057 & 0.0068 & +0.0011 & +18.80\% \\
  assent & 0.0004 & 0.0011 & +0.0007 & +211.56\% \\
     anx & 0.0003 & 0.0009 & +0.0006 & +182.32\% \\
  TenseM & 0.0005 & 0.0009 & +0.0004 & +70.97\% \\
PresentM & 0.0005 & 0.0008 & +0.0003 & +60.84\% \\
\midrule
\multicolumn{5}{c}{\textbf{Largest Decreases}} \\
\midrule
             work & 0.0154 & 0.0145 & -0.0010 & -6.24\% \\
Personal Concerns & 0.0165 & 0.0154 & -0.0010 & -6.23\% \\
           social & 0.0015 & 0.0005 & -0.0010 & -69.05\% \\
           health & 0.0148 & 0.0140 & -0.0008 & -5.11\% \\
              bio & 0.0148 & 0.0140 & -0.0008 & -5.36\% \\
           humans & 0.0007 & 0.0000 & -0.0007 & -98.98\% \\
          relativ & 0.0033 & 0.0026 & -0.0007 & -21.25\% \\
             time & 0.0027 & 0.0021 & -0.0006 & -23.29\% \\
          insight & 0.0008 & 0.0005 & -0.0004 & -45.85\% \\
          discrep & 0.0016 & 0.0013 & -0.0003 & -16.85\% \\
\bottomrule
\end{tabular}
\end{table*}

\end{document}